\documentclass[3p]{elsarticle}
\usepackage{booktabs}
\usepackage{lineno,hyperref}
\modulolinenumbers[5]
\journal{Journal of \LaTeX Templates}
\usepackage{amsmath}
\usepackage{amssymb}
\usepackage{natbib}
\usepackage{amsfonts}
\usepackage{dsfont}
\usepackage{setspace}
\usepackage{subfigure}

\biboptions{authoryear}

\usepackage{color}

\usepackage{comment}
\begin{document}
\onehalfspacing
\begin{frontmatter}

\title{Comparing statistical and machine learning methods for time series forecasting in data-driven logistics -- A simulation study}

\author[1]{Lena Schmid\corref{cor1}}
\ead{lena.schmid@tu-dortmund.de}
\author[2]{Moritz Roidl}
\author[1,3]{Markus Pauly}
\address[1]{Department of Statistics, TU Dortmund University, 44227 Dortmund, Germany}
\address[2]{Chair of Material Handling and Warehousing, TU Dortmund University, 44221
Dortmund, Germany}
\address[3]{UA Ruhr, Research Center Trustworthy Data Science and Security, 44227 Dortmund, Germany}
\cortext[cor1]{Corresponding Author}

\begin{abstract}
 Many planning and decision activities in logistics and supply chain management are based on forecasts of multiple time dependent factors. Therefore, the quality of planning depends on the quality of the forecasts. We compare different state-of-the art forecasting methods in terms of out-of-the-box forecasting performance. Different to most existing research in logistics, we do not do this case-dependent but consider a broad set of simulated time series to give more general recommendations. We therefore simulate various linear and nonlinear time series that reflect different situations.

\end{abstract}

 
\end{frontmatter}

%
 \section{Introduction}
 \label{sec:Into}
%

Forecasting methods are essential for efficient planning in various logistics domains such as warehousing, transport, and supply chain management. They enable companies to anticipate and plan for future demand, capacity needs, and supply chain requirements. Thereby, different logistics applications require different forecasts due to their unique characteristics. 
In the transport domain, e.g., accurate transportation forecasting enables logistics companies to optimize their transportation networks, reduce transportation costs and enhance delivery reliability \citep{huang2020travel, wu2004travel, lin2005review, garrido2000forecasting, wu2021ensemble}. 
 Precise forecasting allows warehouse managers to optimize space use, reduce stock-out risk, and improve overall efficiency \citep{shi2018dynamic, ribeiro2022short}. In supply chain management, accurate forecasts are, e.g., used to optimize resource use across the entire supply chain \citep{feizabadi2022machine,kuhlmann2023dynamic,syntetos2016supply}.
 The above references show that the use of forecasting techniques such as time series models and machine learning methods has become increasingly popular in logistics in recent years. However, there is still a lack of consensus on which method is more effective, especially as most methods of comparison in logistics solely rely on comparing the performance on a few data sets \citep{ensafi2022time, ribeiro2022short}. In fact, different to other fields \citep[e.g.][]{wu2018moleculenet,weber2019essential} there do not exist rigorous benchmark studies in data-driven logistics to the best of our knowledge. 
 In our opinion, the key reason for this is that, outside of specific examples \citep[e.g.][]{niemann2020lara,arora2022analysis}, there is a lack of freely accessible and well-characterized data sets for benchmarking \citep[e.g.][]{reining2019human,awashti2023} in the logistics research domain. This hampers the analysis of domain-specific pros and cons of method choices or the formulation of general recommendations. 
 To overcome this and to be in line with recent recommendations \citep{friedrich2023role}, we therefore focus on simulating data from various statistical time series models that reflect potential logistic scenarios.\\

Time series models have been used in forecasting for several decades and are widely used in logistics for sales or demand forecasting, see e.g. \cite{kuhlmann2023dynamic, shukla2011arima} and the references cited therein. These models are based on historical data and use statistical techniques to identify patterns and trends in the data, which can then be used to make predictions about future demand. Some commonly used time series models in logistics include (seasonal) autoregressive integrated moving averages (ARIMA) and exponential smoothing models. For example, \cite{introduction_ts_sc_02} developed an ARIMA multistage supply chain model that is based on time series models. 
Another example is Prophet \citep{taylor2018forecasting}, a forecasting tool for time series analysis developed by Facebook, which includes additive modeling with components such as seasonality, holidays and trend flexibility. \cite{Prophet} examined ARIMA and Prophet models for predicting supermarket sales. The Prophet models showed superior predictive performance in terms of lower errors. \cite{ExpSmoothing} investigated the performance of double exponential smoothing for inventory forecasting.\\

More recently, machine learning (ML) methods have become increasingly popular for demand forecasting in logistics due to their ability to handle large and complex data sets. There are many literature reviews \citep{scmreview1, scmreview2, scmreview3, scmreview4, scmreview5}, that discuss the use of machine learning techniques in forecasting for supply chain management, including an overview of the various techniques used and their advantages and limitations. However, our comment regarding a lack of neutral benchmarking studies still applies. \\

Several studies have shown that ML methods such as neural networks, support vector regression, and Random Forests can outperform traditional time series models for specific demand forecasting problems. For example, a study by \cite{ensafi2022time} compared the prediction power of more than ten different forecasting models, including classical methods such as ARIMA and ML techniques such as long short-term memory (LSTM) and convolution neural networks, using a single data set containing the sales history of furniture in a retail store. The results showed that the LSTM outperformed the other models in terms of prediction performance. Another study by \cite{kohzadi1996comparison} also compared the forecasting power of ARIMA and neural networks using a single commodity prices data set. Again the neural network performed better than the ARIMA model. Similar results were obtained in \cite{weng2019forecasting} or \cite{siami2018comparison}.
However, other studies have found mixed results, with some suggesting that time series models perform better than ML methods. For instance, \cite{palomares2009arima} compared the forecasting accuracy of ARIMA and neural network models in predicting wind speed for short time intervals. The results showed that the performance of both can be very similar, indicating that a more simple and interpretable forecasting model could be used to administrate energy sources. A comparison of daily hotel demand forecasting performance of SARIMAX, GARCH and neutral networks also showed that both time series approaches outperformed the neural networks \citep{ampountolas2021modeling}. In the latter examples, one reason may also be the difficulty of tuning complex machine learning procedures. That's one reason why we focus on out-of-the-box machine learning methods in our study.\\

The comparison of the forecasting performance of ML methods and time series models in logistics has significant implications for businesses seeking to improve their forecasting accuracy. By identifying the most effective forecasting methods, businesses can make better-informed decisions about production, inventory management, and resource allocation.
Thus, this work aims to provide a comprehensive comparison of the forecasting performance of time series models and ML methods. Different from the above-mentioned works that merely focus on single use cases, this task needs more variation in the data sets under study. To this end, we compare various forecasting methods in terms of out-of-the-box forecasting performance on a broad set of simulated time series. We thereby simulate various linear and nonlinear time series that are of importance for logistics and study the one-step forecast performance of different statistical learning methods.\\

This work is structured as follows: Section~\ref{sec:Methods} presents the different used forecasting methods. More precisely, the (seasonal) ARIMA and TBATS models are presented. In addition, the machine learning approaches (Random Forest and XGBoost) are described in more detail. Section~\ref{sec:Design} presents the simulation design and framework, while Section~\ref{sec:Results} summarizes the main simulation results. In Section 5, an illustrative real-world data example is analyzed before the manuscript concludes with a discussion of our findings and an outlook for future research (Section~\ref{sec:Conclusion}).
%
 \section{Methods}
 \label{sec:Methods}
%
In this section, we explain the one-step forecasting methods under investigation. There are various strategies for modeling and forecasting time series. Traditional time series models, including moving averages and exponential smoothing, follow a linear approach in which the predictions of future values are linear functions of past observations. Due to their relative simplicity in terms of understanding and implementation, linear models have found application in many forecasting problems \citep{fan2021well, nyoni2018modeling, benvenuto2020application}. To overcome the limitations of linear models and account for certain nonlinear patterns observed in real-world problems, several classes of nonlinear models have been proposed in the literature. Examples cover the threshold autoregressive model (TAR) \citep{tsay1989testing} or the generalized autoregressive conditional heteroscedastic model (GARCH) \citep{francq2019garch}. Although some improvements have been noted, the utility of their application to general prediction problems is limited \citep{de1992some}: Since these models were developed for specific nonlinear patterns, they are often unable to model other types of nonlinearities. Here, machine learning methods have been proposed as an alternative for time series forecasting \citep{bontempi2012machine, ahmed2010empirical}. Since it is impossible to cover the entire spectrum of machine learning models and time series methods in our simulation study, we limit ourselves to a selection of what we consider the most common algorithms in data-driven logistics. 
To evaluate the performance, we compare these methods with a naive approach, where the last observation of the time series is used as a prediction. The time series (Subsection~\ref{sec:tsmodels}) and machine learning methods (Subsection~\ref{sec:mlmodels}) under study are explained in more detail in the next two subsections.

\subsection{Time Series Methods}\label{sec:tsmodels}
We focus on three different time series models: ARIMA, SARIMA, and TBATS. The first two models are among the most popular models in traditional time series forecasting \citep{brockwell2002introduction, hyndman2018forecasting} and are often used as benchmark models for comparison with machine learning algorithms \citep{al1999artificial, zhang2001simulation, hwarng2001insights}. In addition, TBATS models combine many different approaches that are commonly used in forecasting.

\paragraph{ARIMA}
Autoregressive integrated moving average (ARIMA)\citep{box2105} model is a generalized model of the autoregressive moving average (ARMA) model and builds a composite model of the time series \citep{shumway2000time}. Denoted as ARIMA($p$, $d$, $q$), $p, q, d \in \mathbf{N},$ the model is characterized by three key components:
\begin{itemize}
 \item AR (Autoregression): Represents the regression of the time series on its own past values, capturing dependencies through lagged observations. The number of lagged observations included in the models is given by $p$.

 \item I (Integrated): The differencing order ($d$) indicates the number of times the time series is differenced to achieve stationarity. This transformation involves subtracting the current observation from its $d$-th lag, which is crucial for stabilizing the mean and addressing trends.

 \item MA (Moving Average): Incorporates a moving average model to account for dependencies between observations and the residual errors of the lagged observations ($q$).
 \end{itemize}
In general a time series $\{x_t\}_t$ generated from an ARIMA($p$, $d$,$q)$ model has the form
\begin{equation*}
 \sum_{i=1}^p \phi_i \Delta^d x_{t-i} = \sum_{j=0}^q \theta_j \varepsilon_{t-j}, 
\end{equation*}
where $p, d, q \in \mathbb{N}$, $\phi_1, \ldots, \phi_p \in \mathbb{R}$ are the autoregressive coefficients, $\theta_1, \ldots \theta_q \in \mathbb{R}$ are the moving average coefficients and $\varepsilon_t$ denotes the residuals or the errors at time $t$. The residuals are often assumed to follow a white noise process, represented by a sequence of uncorrelated random variables with zero mean and finite second moment. The difference operator $\Delta$ is defined as
$\Delta: \mathbb{R} \rightarrow \mathbb{R}$ with $x_t \rightarrow x_t -x_{t-1}.$

\paragraph{SARIMA}
With seasonal time series data, short-term non-seasonal components likely contribute to the model. Therefore, we need to estimate a seasonal ARIMA model incorporating non-seasonal and seasonal factors into a multiplicative model \citep{shumway2000time}. The general form of a seasonal ARIMA model is denoted as SARIMA$(p, d, q)(P, D, Q)_m$, where $p$ is the non-seasonal AR order, $d$ is the non-seasonal differentiation, $q$ is the non-seasonal MA order, $P$, $D$ and $Q$ are the similar parameters for the seasonal part. The parameter $m$ denotes the number of time steps for a single period.

\paragraph{TBATS}
For time series data exhibiting complex and diverse seasonal patterns, TBATS (Trigonometric Seasonal Exponential Smoothing) is a robust modeling approach. Introduced as an extension of exponential smoothing methods, TBATS accounts for different seasonalities through a combination of trigonometric functions and exponential smoothing \citep{de2011forecasting}. The model is particularly effective in handling multiple seasonal cycles, making it suitable for data sets with intricate temporal structures.

The general form of a TBATS model consists of several components as described below:
\begin{itemize}
 \item {T (Trend):} Captures the overall trend in the time series using an exponential smoothing mechanism.
 
 \item {B (Box-Cox Transformation):} Applies the Box-Cox transformation \citep{box1964analysis} to stabilize variance and ensure the homogeneity of variances.
 
 \item {A (ARIMA Errors):} Incorporates ARIMA errors to capture any remaining non-seasonal dependencies.
 
 \item {S (Seasonal):} Utilizes trigonometric functions to model multiple seasonal components, accommodating various seasonal patterns.
\end{itemize}

\subsection{Machine Learning Methods}\label{sec:mlmodels}
Machine learning methods are increasingly being used to address time series prediction problems. In fact, there exist too many approaches to consider in a comparison study like ours. We therefore restricted ourselves to a class that has already been successfully used for predictions in the logistics context \citep{ji2019application, islam2020prediction, ma2018estimating, huang2020travel, kuhlmann2023rodd}: Tree-based ensemble learners. We thereby focus on two models, each studied with and without differencing: Random Forest and XGBoost on trees which are briefly introduced below. 

 \paragraph{XGBoost}
 Gradient boosting is an ensemble machine learning technique often used in classification and regression problems and is particularly popular in predictive scenarios \citep{aguilar2021short}. As an ensemble technique, gradient boosting combines the results of several weak learners, referred to as base learners, with the aim of building a model that generally performs better than the conventional single machine learning models. Typically, gradient boosting utilizes decision trees as base learners. Like other boosting methods, the core idea of gradient boosting is that during the learning procedure, new models are built and fitted consecutively and not independently to provide better predictions of the output variable. Thereby, new base learners are constructed with the aim of minimizing a loss function associated with the whole ensemble. Instances that are not predicted correctly in previous steps and score higher errors are correlated with larger weight values so that the model can focus on them and learn from its mistakes.\\

XGBoost stands for Extreme Gradient Boosting and is a specific implementation of gradient boosting \citep{chen2016}. It incorporates randomization and regularization techniques to reduce overfitting while increasing training speed. Moreover, it computes second-order gradients of the loss function, which provides more information about the gradient’s direction, making it easier to minimize the loss function. \\

In general, the hyperparameters for XGBoost can be divided into two categories \citep{chen2016}: General boosting parameters, including the number of iterations and the learning rate, which controls how much information from a new tree will be used in the boosting step. Second, in base learner dependent parameters. When trees are used as base learners, the additional hyperparameters are used to control the complexity of the individual trees. Examples include limiting the maximum tree depth or specifying a minimum number of samples in each leaf \citep{therneau1997}.
There also exists other boosting variants \citep{schapire2013boosting, friedman2002stochastic, mayr2014evolution}, but we concentrate on XGBoost as it has emerged as one of the key machine learning models for prediction and was also referred to as \textit{`the Queen of Machine Learning'} \citep{queen} in this context. XGBoost models have also been used for time series forecasting, e.g., \cite{luo2021time, alim2020comparison}. For example, in \cite{zhang2021time} the potential of XGBoost in retail for predicting store sales was investigated while \citep{huang2020travel} studied this for predicting the travel time of NYC cabs.
 
\paragraph{Random Forest}
A Random Forest \citep{Breiman2001} is a machine learning method based on
building ensembles of decision trees. It was developed to address predictive shortcomings
of traditional Classification and Regression Trees (CARTs) \citep{breiman2017classification}. Random Forests consist of a large number
of weak decision tree learners, which are grown in parallel to reduce the bias and variance
of the model at the same time \citep{Breiman2001}. For training a Random Forest, 
bootstrap samples are drawn from the training data set. Each bootstrap sample is then used to grow a(n unpruned) tree. Instead of using all available features in this step,
only a small and fixed number of randomly sampled $m_{try}$ features are selected as split
candidates. A split is chosen by the CART-split criterion for regression, i.e., by minimizing the sum of squared errors in both child nodes. Instead of the CART-split criterion, many
other distances, such as the least absolute deviations of the mean (L1-norm), can also be used. These steps are repeated until $B$ such trees are grown, and new data is predicted by taking the mean of all $B$ tree predictions.
The most important hyperparameters for the Random Forest \citep{ranger} are:
\begin{itemize}
 \item $B$ as the number of grown trees. Note that this parameter is usually not tuned since it is known that more trees are better.
 \item The cardinality of the sample of features at every node is $m_{try}$.
 \item The minimum number of observations that each terminal node should contain (stopping criteria).
\end{itemize}
Though there exist other variants of bagged tree-based ensembles \citep{geurts2006extremely,goehry2023random}, we concentrate on the Random Forest as it is the best known method that is often seen as the machine learning benchmark procedure \citep[e.g.][]{portoles2018electricity}. In addition, Random Forests have also been frequently used for time series forecasting \citep{huang2020travel, kane2014comparison}. For example, in \cite{salari2022real}, a Random Forest approach was used to model real-time delivery time forecasts in online retailing while \cite{vairagade2019demand} applied Random Forest to predict product demand for grocery items. 

While machine learning methods are quite en vogue, we should not neglect the advantages of time series methods in terms of interpretability. Here, time series approaches enable a clearer understanding of the factors influencing the predictions.

%
\section{Simulation Set-up}
\label{sec:Design}
%
In our simulation study, we compare the one-step forecast prediction performance of the methods described in Section 2. All simulations were conducted in the statistical computing software \texttt{R} \citep{R}. We use the \texttt{forecast} package \citep{forecast} for all time series approaches under consideration. For the machine learning methods, we used the \texttt{ranger} \citep{ranger} and \texttt{xgboost}\citep{xgboost} packages for Random Forest and XGBoost, respectively.
The concrete simulation settings and data generating processes (DGPs) are described below. 
\paragraph{Data Generating Processes}
We consider twelve DGPs in total - an autoregressive model (AR), two bilinear models (BL), two nonlinear autoregressive models (NAR), a nonlinear moving average model (NMA), two sign autoregressive models (SAR), two smooth transition autoregressive models (STAR) and two TAR models. They are summarized in Table~\ref{tab:dgps}, where the error terms $\varepsilon_t$ are independent and identically distributed with a standard normal distribution. 
\renewcommand*{\arraystretch}{2}
\begin{table}[h]
 \centering 
 \caption{Data generating processes (DGPs) used in the simulation study. The error terms $\varepsilon_t$ are i.i.d $\mathcal{N}(0,1)$.}
 \begin{tabular}{|l|l|l|}
 \hline
 Model Type & Variant(s)& Data generating process\\\hline
 Autoregressive& AR& $x_t=0.5x_{t-1}+0.45x_{t-2}+\varepsilon_t$,\\
 Bilinear& BL 1& $x_t=0.7 x_{t-1}\cdot\varepsilon_{t-2}+\varepsilon_t$, \\
 & BL2& $x_t=0.4 x_{t-1}-0.3x_{t-2}+0.5x_{t-2}\cdot\varepsilon_{t-1}+\varepsilon_t$,\\
 Nonlinear Autoregressive&NAR 1& $x_t=\tfrac{0.7|x_{t-1}|}{|x_{t-1}|+2}+\varepsilon$,\\
 &NAR2 & $x_t=\tfrac{0.7|x_{t-1}|}{|x_{t-1}|+2}+\tfrac{0.35|x_{t-2}|}{|x_{t-2}|+2}+\varepsilon$,\\
 Nonlinear Moving Average& NMA&$x_t=\varepsilon_t-0.3\varepsilon_{t-1}+0.2\varepsilon_{t-2}+0.4\varepsilon_{t-1}\varepsilon_{t-2}-0.25\varepsilon_{t-2}^2$,\\
 Sign Autoregressive &SAR 1& $x_t=\text{sign}(x_{t-1})+\varepsilon_t$,\\
 &SAR 2& $x_t=\text{sign}(x_{t-1}+x_{t-2})+\varepsilon_t$,\\
 Smooth Transition& STAR 1& $x_t=0.8\varepsilon_t-\frac{0.8\varepsilon_{t-1}}{1+\exp(-10x_{t-1})}+\varepsilon_t$,\\
 Autoregressive&STAR 2& $x_t=0.3x_t+0.6x_{t-2}+\frac{0.1-0.9x_{t-1}+0.8x_{t-2}}{1+\exp(-10x_{t-1})}+\varepsilon_t$,\\
 Threshold Autoregressive&TAR 1& $x_t=\left\{\begin{array}{cc}
 0.9x_{t-1}+\varepsilon_t & \text{ if }|x_{t-1}|\leq 1\\
 -0.3x_{t-1}-\varepsilon_t & \text{ if }|x_{t-1}|>1
 \end{array}\right.$\\
 &TAR 2& $x_t=\left\{\begin{array}{cc}
 0.9x_{t-1}+0.05x_{t-2}+\varepsilon_t & \text{ if }|x_{t-1}|\leq 1\\
 -0.3x_{t-1}+0.65x_{t-2}-\varepsilon_t & \text{ if }|x_{t-1}|>1
 .\end{array}\right. $\\\hline
 \end{tabular}
\label{tab:dgps}
\end{table}

Similar models have been used to evaluate time series forecasts \citep{ZHANG1} and are of importance in data-driven logistics. In particular, autoregressive models (AR, NAR1, NAR2) were chosen to capture the persistence observed in historical logistics demand \citep{luong2007measure}. Bilinear models (BL1, BL2) reflect the complex interactions within logistics networks where different components contribute to the observed patterns. The non-linear moving average (NMA) model is suitable for scenarios with complex interdependencies between multiple factors. Sign autoregressive models (SAR1, SAR2) are suitable for situations in which events or conditions have a directional influence on future events. Smooth transition autoregressive models (STAR1, STAR2) mimic logistics systems where demand changes gradually due to external factors \citep{ubilava2012modeling} and threshold autoregressive models (TAR1, TAR2) represent logistics processes with different regimes based on specific conditions \citep{ricky2005threshold}. This diverse set of DGPs depicts many aspects of the multi-layered nature of logistics data, which includes persistence, interactions, complicated dependencies, directional influences, smooth transitions and different regimes. In the absence of comprehensive benchmark problems, this set-up allows us to evaluate the adaptability of forecasting methods in dynamic logistics scenarios. 

\paragraph{Additional Complexities} To add additional complexity to the analysis, we have incorporated settings with a jump process and a random walk \citep{shumway2000time} into each DGP. The jump process introduces sudden regime changes (which may occur in logistics due to unforeseeable events), while the random walk adds noise to the data (which may occur in logistics settings with increased complexity or less accurate measurements). Thus, our study considers four different scenarios: (1) the DGP without additional complexity, (2) the DGP superposed with the jump process, (3) the DGP superposed with random noise, and (4) the DGP superposed with both the jump process and random noise. The \textit{jumps} are modeled using a compound Poisson process $\{p_t\}_t$ \citep{kingman1992poisson}. The original DGP $\{x_t\}_t$ is then superposed by ${p_t}$ as follows
$$x_t^*=x_t+p_t,$$
where $x_t^*$ denotes the resulting DGP, and the compound Poisson process is given by
$$ p_t=\sum_{i=1}^{N_t} Z_i,$$
where $N_t$ follows a Poisson distribution with parameter $\lambda$ and $Z_i \sim \mathcal{N}(0,\sigma_p^2).$ For the jump experiments we set $\sigma_p^2$ to 1. A larger $\sigma_p^2$ results in larger jumps in magnitude, while the mean over positive and negative jumps remains zero. The parameter $\lambda$ is set to $\frac{n}{10}$, where $n$ denotes the length of the generated time series. This means that, on average, a jump is expected to occur after every $\lambda$ period.
Superposing the DGP with the compound Poisson process results in a mean shift by the actual jump size that occurred at each jump event. As mentioned before, the noise is modeled by a \textit{random walk} $\{w_t\}_t$ with
$$w_t= w_{t-1}+e_t,$$
where $e_t\sim \mathcal{N}(0, \sigma_{rw}^2)$. In our study, we choose $\sigma_{rw}^2$ in such a way that we obtain a setting with medium noise, i.e., a signal-to-noise ratio (SNR) of four. The SNR \citep{box1988signal} is a measure that characterizes the strength of the signal relative to the background noise. A higher SNR indicates a clearer and more discernible signal amidst the noise. By including the random walk, we achieve a resulting DGP that is globally nonstationary due to the random walk overlay.\\

\paragraph{Additional Queueing Models} 
Beyond these 48 simulation models, we include the M/M/1 and M/M/2 queueing models \citep{queueing} in our study. Queueing models are commonly used in logistics, operations research and industrial engineering to study the behavior of waiting lines or queues \citep{queueing1, queueing2, queueing3, queueing4}. Both models have numerous real-world applications, such as in call centers \citep{brown2005statistical}, healthcare facilities \citep{green2006queueing}, and transportation systems \citep{radmilovic1996some}. The M/M/1 model is a classic queueing model that assumes a single queue and one server. It is a stochastic model, where customer arrivals are assumed to follow a Poisson process, and service times are exponentially distributed. The M/M/1 model can be used to analyze the expected waiting time, the number of customers in the queue, and the expected server utilization.
The M/M/2 model is a variation of the M/M/1 model that assumes two parallel servers. According to \cite{queueing4}, we set the arrival rate to four and the service rate to two. 

\paragraph{Number of different Settings} For each setting, we generate time series of length $n$ from the respective DGPs with $n\in\{100, 500, 1000\}.$ In total, this results in $150$ (= $12$ (time series DGPs) $\times\,4$ (further complexity)$+ 2$ queueing models) $\times\,3$ (lengths)) different simulation settings for each forecasting method.

\paragraph{Data Preprocessing}
To forecast time series using a machine learning algorithm, the sliding window approach \citep{dietterich2002machine} is used. In this approach, a fixed-sized window is moved over the time series data, and at each step, the data within the window is used as input to a machine learning algorithm for prediction. One advantage of the sliding window approach is that it allows the machine learning algorithm to capture the temporal dependencies and patterns in the data. The window size is an important parameter in this approach \citep{savva2020effects}. If the window size is too small, it may not capture the relevant information in the data, while if it is too large, it may introduce unnecessary noise and reduce the accuracy of the model. We consider sliding window sizes of 2, 4, 8 and 16 and study which size is best suited for the different time series lengths 100, 500 and 1000. Furthermore, in machine learning-based time series forecasting, we explore two approaches: one using the original time series and the other using the differentiated time series as input. The latter is essential as trees cannot forecast outside the range observed so far and to enhance stationarity in the time series.

\paragraph{Choice of Parameters}
In order not to have to discuss the different possibilities for hyperparameter tuning of the machine learning algorithm, we use the default values recommended in the literature \citep{Breiman2001, ranger,Hastie2009}. This has the additional advantage of a reduced runtime. Thus, each ensemble learner consists of $500$ trees, the inner bootstrap sample is equal to $m_{try}=\lfloor \tfrac{p}{3}\rfloor$, where $p$ denotes the number of features, the number of sample points in the bagging step is equal to the sample size. Each terminal node should at least contain five observations. For XGBoost, we use a learning rate of 0.3 and a maximal depth of 6. To estimate the parameters of the time series approaches, we use the algorithms implemented in the R-package \texttt{forecast}.

 \paragraph{Evaluation Measure}
Since the mean square error (MSE) and the mean absolute percentage error (MAPE) are widely used in the forecasting of time series in logistics \citep{kuhlmann2023dynamic}, we use them as evaluation measures, which are calculated over 1,000 repeated forecasting steps. The MSE measures the model's accuracy, expressed as the average squared difference between observed and predicted values. Simultaneously, the MAPE, calculated as the average percentage difference between observed and predicted values, offers insights into the model's relative performance.

%
 \section{Results}
 \label{sec:Results}
%
In this section, we describe the results of the simulation study. In particular, we present the MSE
of the different forecasting algorithms under various simulation configurations. The analysis of the MAPE results can be found in the Appendix. We start with the performance of the methods for queueing models.
\subsection{Predictive Power in Queueing Models}

\begin{figure}[h!]
 \centering
\includegraphics[scale=0.57]{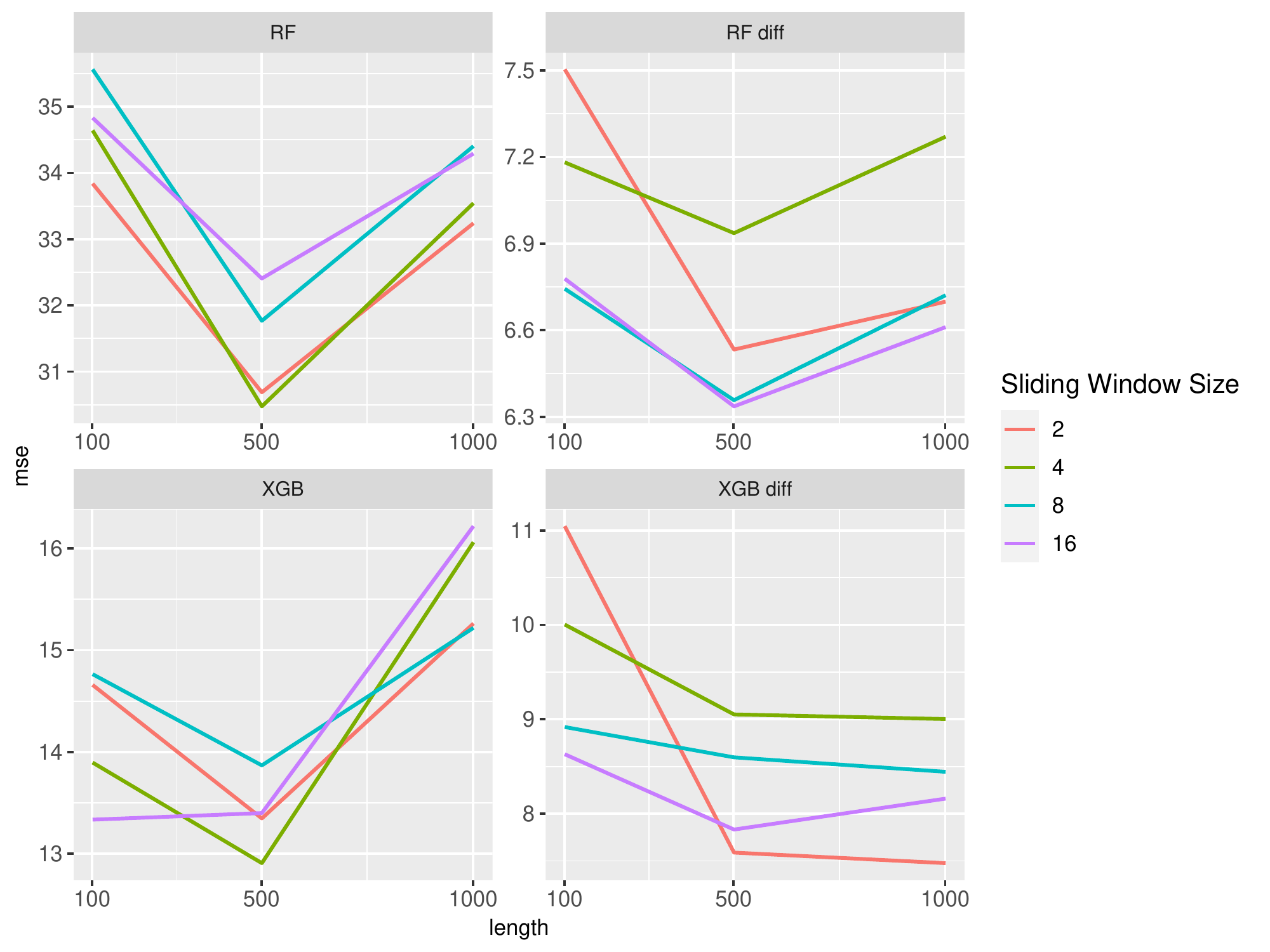}
 \caption{MSE of ML approaches separated by the sliding window size for the M/M/1 setting. XGB stands for XGBoost and RF for Random Forest; diff in the method name indicates that the data were differentiated.}%
 \label{fig:MM1Inp}
\end{figure}

\begin{figure}[h!]
 \centering
\includegraphics[scale=0.57
]{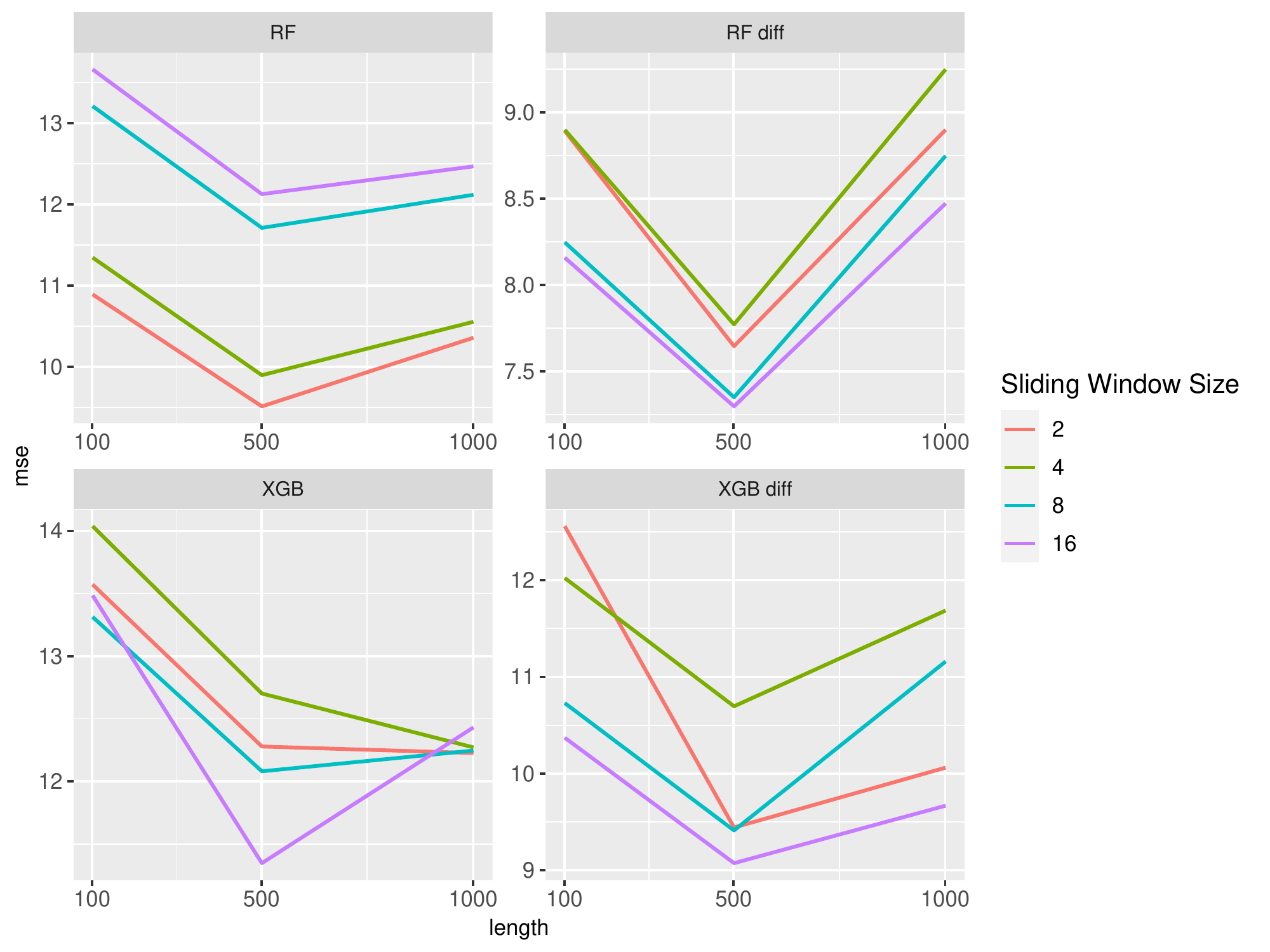}
 \caption{MSE of ML approaches separated by the sliding window size for the M/M/2 setting. XGB stands for XGBoost and RF for Random Forest; diff in the method name indicates that the data were differentiated.}%
 \label{fig:MMInp}
\end{figure}

The influence of the different sliding window sizes and the differentiation is shown in Figures \ref{fig:MM1Inp} and \ref{fig:MMInp}.
Generally, differentiation improves the prediction power of both ML approaches in both settings. Especially for the Random Forest, the MSE decreases by one-fifth after differentiation. The lengths of the time series only have a minor influence on the MSE. The Random Forest with differentiated data outperformed the other methods for all lengths. Comparing the effects of sliding window sizes, we find slight differences in performance. Random Forests have smaller MSE values with smaller sliding windows in both settings, while larger window sizes slightly improve performance in the other approaches. \\

\begin{figure}[h!]
 \centering
\includegraphics[scale=0.55]{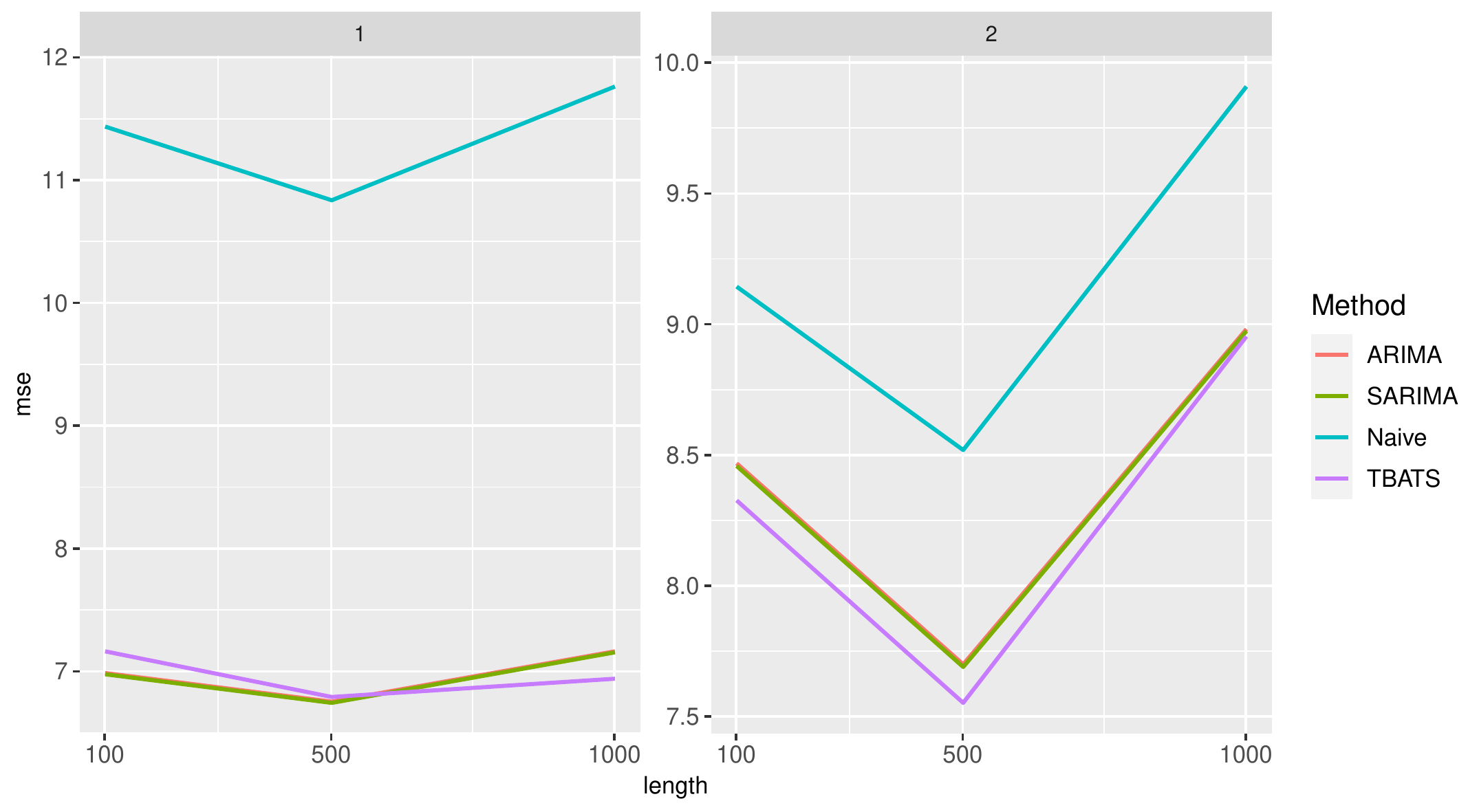}
 \caption{MSE of time series and naive approaches for the M/M/1 (left) and M/M/2 (right) setting. ARIMA and SARIMA models have identical MSE values, as no seasonality was present.}%
 \label{fig:MMTS}
\end{figure}

The predictive power of the time series and naive approaches are given in Figure \ref{fig:MMTS}. Note that both ARIMA and SARIMA models have identical MSE values. In both cases, the time series approach performs better than the naive approach. However, the difference in performance is smaller for M/M/2. Again, the influence of the time series length is marginal. While all time series approaches perform similarly in the M/M/1 setting, the TBATS method has slightly smaller values in the M/M/2 setting.\\

 In both scenarios, the Random Forest approach with differenced data consistently showed the smallest MSE. However, the differences between this method and the time series approaches were not great.

\subsection{Predictive Power in the Different Time Series Settings}
\begin{figure}[h!]
 \centering
 \includegraphics[width=0.79\textwidth]{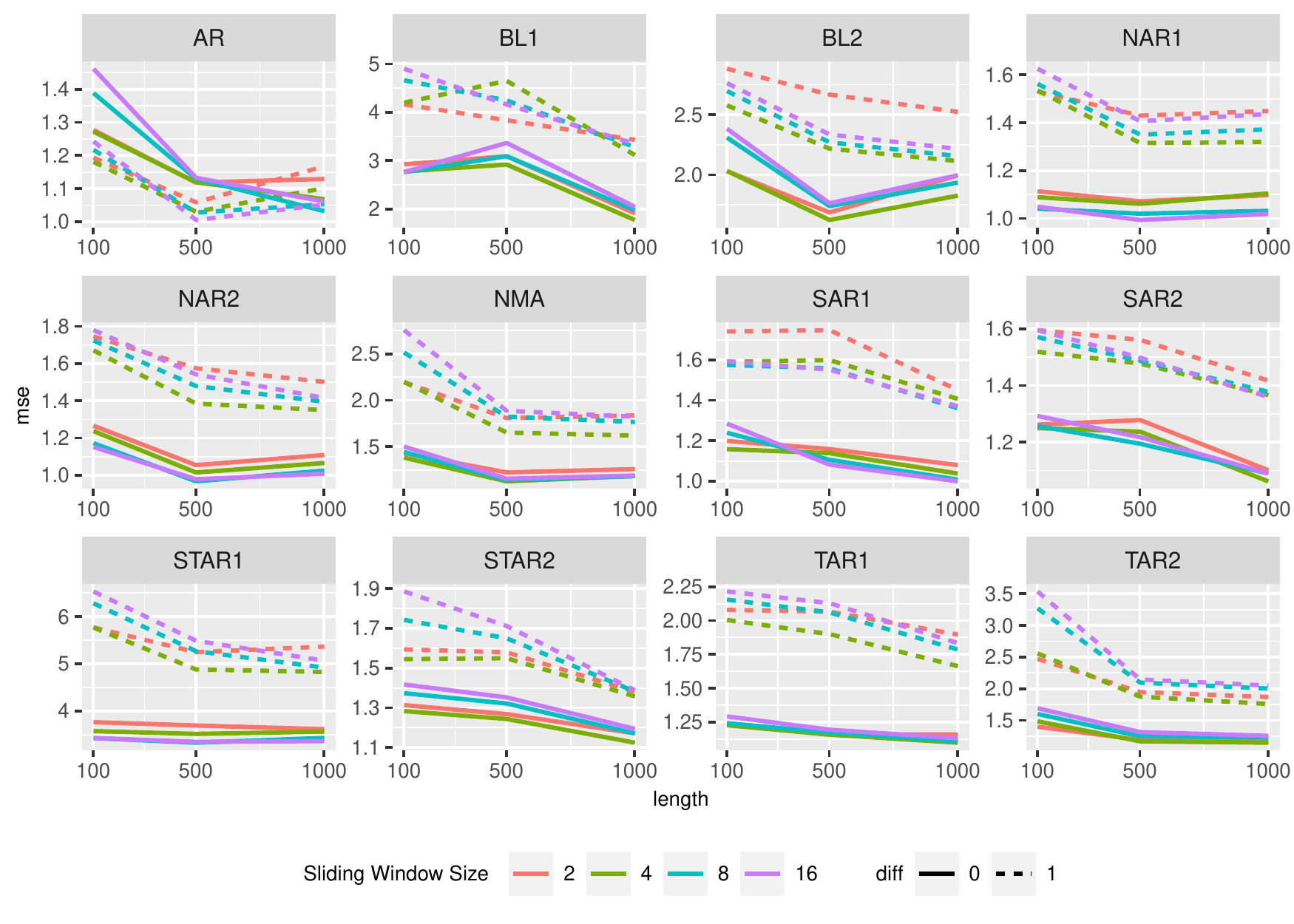}
 \caption{MSE of the Random Forest approaches separated by the sliding window size and differentiation for the different data generating processes.}
 \label{fig:RF}
\end{figure}

In the following, we analyze the performance of the methods for the DGPs described in Table~\ref{tab:dgps}.
When comparing the influence of sliding window size and differentiation on the performance 
of Random Forest across all settings (Figure~\ref{fig:RF}),
we observed that non-differentiation resulted in smaller MSE values except for the AR setting. In the AR setting, differentiation slightly outperformed non-differentiation. However, it should be noted that as the length of the time series increases, the differences between the two approaches become negligible. In all settings, the MSE values slightly decrease with an increase in time series length. The sliding window size has a small influence on the prediction power and shows similar behavior across different time series lengths.\\
 \begin{figure}[h!]
 \centering
 \includegraphics[width=0.79\textwidth]{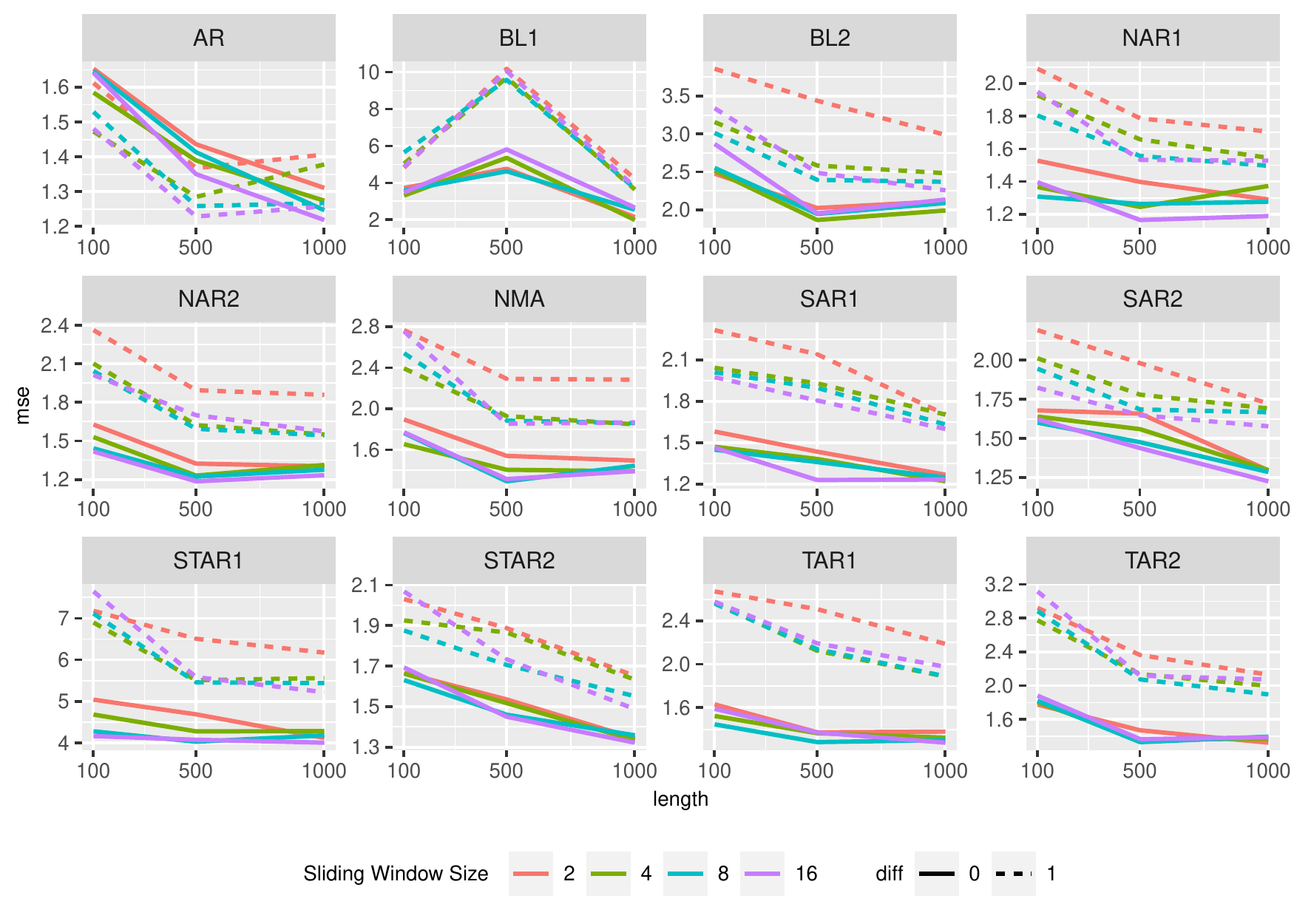}
 \caption{MSE of XGBoost approaches separated by the sliding window size and differentiation for the different data generating processes.}
 \label{fig:XGBoost}
\end{figure}

 Similar observations can be made for XGBoost, see Figure \ref{fig:XGBoost}. The sliding window's size and the time series length have a small effect on the performance quality. For all DGPs, the MSE values decrease slightly with increasing time series length, except for BL1. Here, the MSE values first increase. The XGBoost approaches generally have slightly larger MSE values than the Random Forest approaches. \\

\begin{figure}[h]
 \centering
 \includegraphics[width=0.8\textwidth]{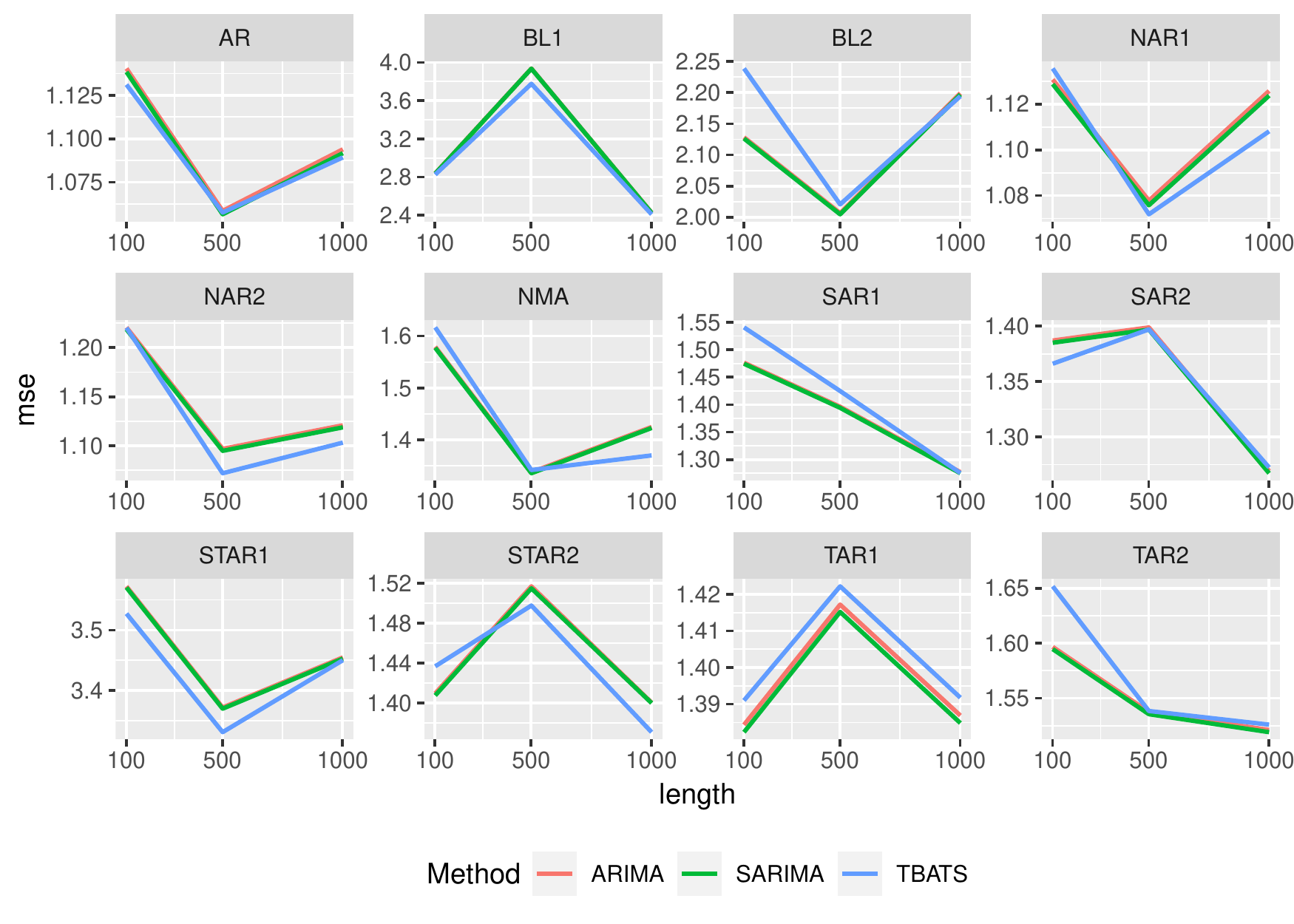}
 \caption{MSE of the time series approaches for the different data generating processes.}
 \label{fig:TS}
 \end{figure}

 Figure \ref{fig:TS} shows the MSE values for the time series approaches. The performance of the time series approaches is comparable to that of the Random Forest. All methods have very similar MSE values. The time series length has only a minor impact on the predictive power, except for the BL1 setting. As observed for the XGBoost approaches, MSE values in this setting first increase and then decrease with increasing time series length. \\

 Additional results can be found in the Appendix. Figure \ref{fig:naive} therein, e.g., shows that the naive approach exhibits the largest MSE values compared to all methods. 
 Thereby the performance of the naive approach is dependent on the DGP and the length of the time series. For BL2, longer time series lengths generally lead to better performance, but for NAR1 the performance may slightly decrease. For AR, BL1, and NMA models, the MSE values typically decrease initially and then slightly increase as the time series length increases. Conversely, NAR2, SAR1, SAR2, STAR1, STAR2, TAR1 and TAR2 tend to show the opposite trend.

\subsection{Influence of the Additional Complexities on the Predictive Power}
Based on the findings of the previous sections, we focus on the simulation results obtained with a sliding window size of 8, as the choice of this size is due to the consistent performance observed with different sizes. Details of the results with other window sizes can be found in the Appendix, but a moderate size of 8 balances computational efficiency and information incorporation. Below we first consider the influence of an additional jump process before discussing the white noise results.\\

\begin{figure}[h!]
 \centering
 \includegraphics[width=0.65\textwidth]{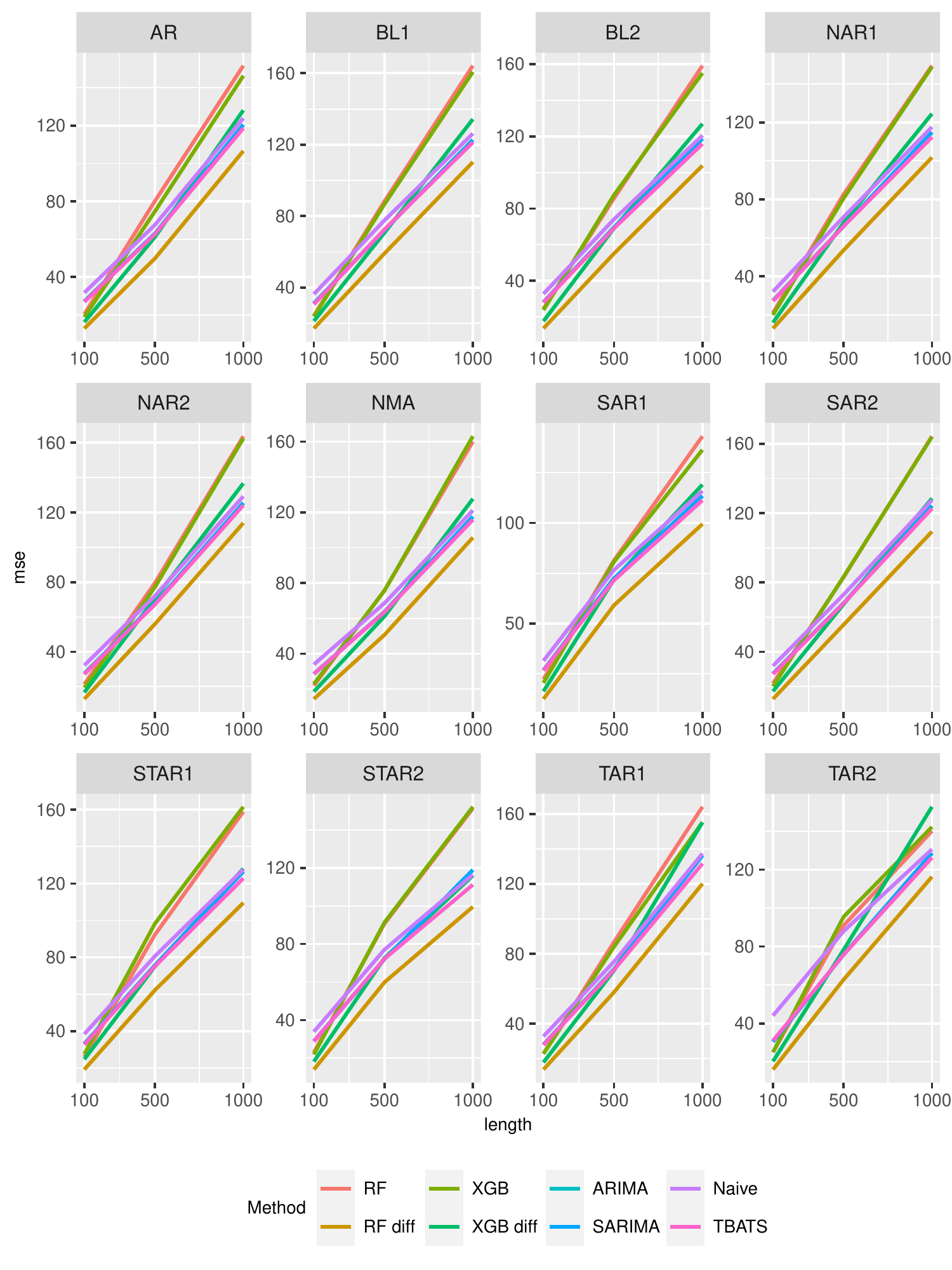}
 \caption{MSE values of all methods and data generating processes superposed by a compound Poisson process.}
 \label{fig:PP}
\end{figure}

The influence of the jumping process can be seen in Figure~\ref{fig:PP}. All MSE values increase monotonically with increasing sample size, indicating that the jump process significantly impacts predictive performance. Note that as time series length increases, the Random Forest approach with differentiated data outperforms all other approaches. Using the differenced data significantly improves the MSE values for both ML approaches, particularly for 
increasing time series length. The predictive performance of the time series approaches is similar for all DGPs and slightly better than that of the naive approach.\\\

\begin{figure}[ht!]
 \centering
 \includegraphics[width=0.65\textwidth]{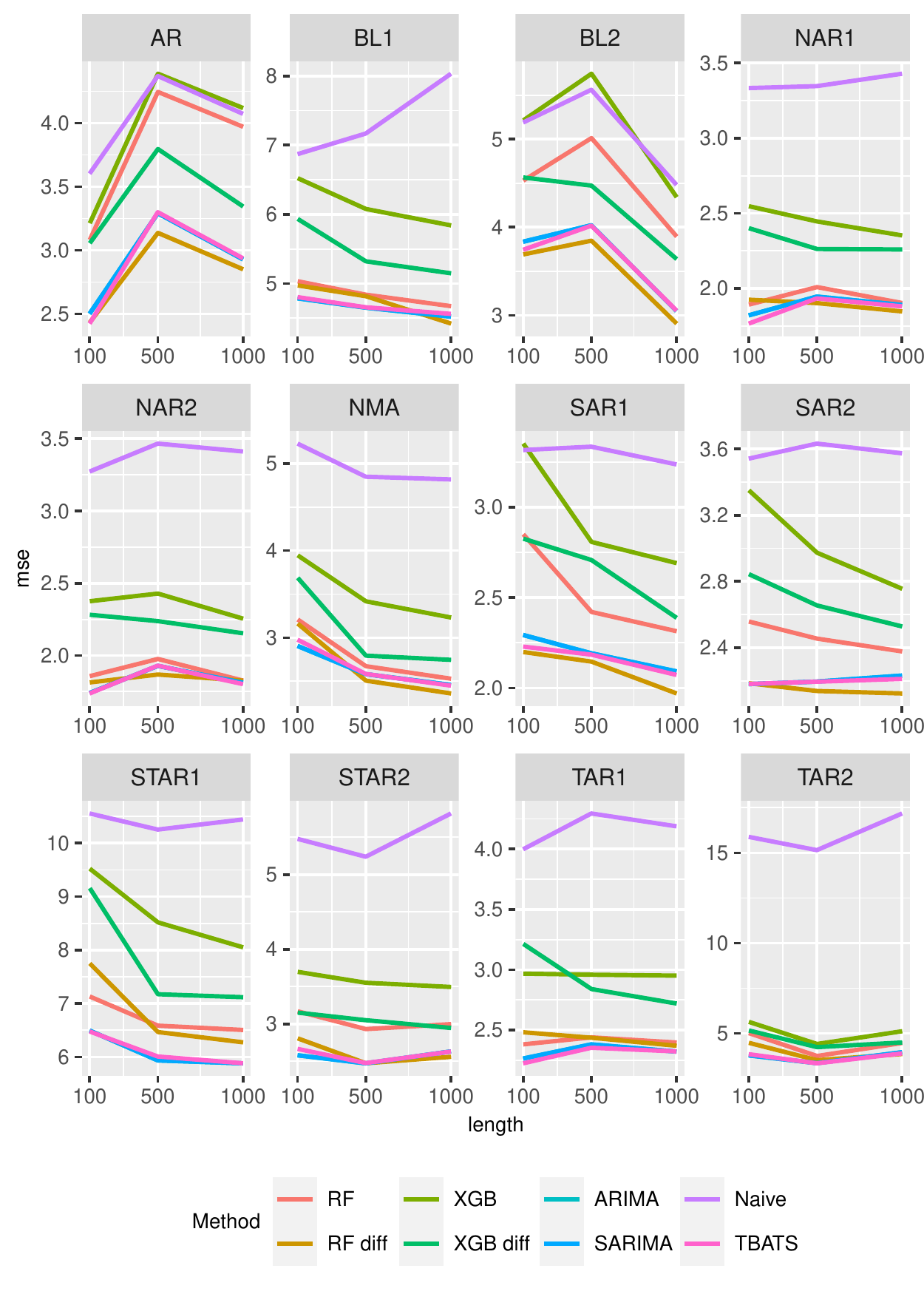}
 \caption{MSE values of all methods and data generating processes superposed by a random walk.}
 \label{fig:RW}
\end{figure}

Figure~\ref{fig:RW} summarizes the prediction results for all methods and all DGPs superposed by a random walk.
Here, the time series length has only a minor influence on the prediction performance of the data overlaid with a random walk. For the AR and BL2 settings, the MSE values increase slightly when the time series length is increased from 100 to 500. For all other DGPs, the MSE values decrease slightly, except for the naive approach. The naive approach has the highest MSE values for all settings, followed by XGBoost, except for BL2. Here, both approaches have similar values. The performance of the other methods depends on the respective setting.\\

For the settings, AR, BL2, SAR1 and SAR2, Random Forest with differenced data again shows the smallest MSE values, while the time series approaches show slightly larger values. Note that the XGBoosts with differentiated data perform better in these settings than the Random Forests with non-differentiated data. In the BL1, NAR1, NAR2, NMA and STAR2 settings, only minor differences in the performance of the Random Forests and time series approaches can be observed. When comparing the two XGBoost approaches in these settings, the differentiation reduces the MSE. The ML approaches show larger MSE values in the STAR1, TAR1 and TAR2 settings than the time series approaches, with Random Forests performing better than the XGBoost method.\\

The influence of both complexities, the random walk and the Poisson process, on the prediction performance is shown in Figure~\ref{fig:both} in the Appendix. Similar to the case where a composite Poisson process is superposed on the data, we observe an increase in MSE values with increasing time series length for all settings. In particular, for time series lengths of 500, we obtain MSE values of more than 2,000.

\subsection{Summarizing all Results}

To evaluate the prediction performance across the spectrum of simulation settings, we calculate the median rank for each prediction method in Table~\ref{tab:summary}. The ranking is based on the MSE values, with rank~1 indicating the method with the lowest MSE. Each entry in the table represents the median rank of a particular prediction method in all settings of a particular DGP model described in Section 2. Furthermore, the results for the ranking take into account the performance of machine learning algorithms with a sliding window size of 8.

\renewcommand{\arraystretch}{1.1} 
\begin{table}[h]
 \centering
 \caption{Median performance rank of forecasting methods across different simulation settings and different time series lengths. 
 Rankings are based on MSE values, with rank 1 indicating the method with the lowest MSE.}
\begin{tabular}{|l|c|c|c|c|c|}
\hline
Method&Queueing &\multicolumn{4}{c|}{DGPs (Table 1) with} \\
&models& no add. complexity&jump& random walk & both\\
\hline
Random Forest &7 &1 & 7&5 &7 \\
 Random Forest Diff& 1&6 &1 & 1&1 \\
 XGBoost& 7&5 & 7& 7& 7\\
 XGBoost Diff&5 &7 & 5&6 &6 \\
 ARIMA& 2.5& 3& 3& 3& 3\\
 SARIMA&2.5 &3 & 3&3 & 3\\
 TBATS& 3& 3&3 &3 & 3\\
 Naive&6 & 8& 6& 8& 5\\
 \hline
\end{tabular}
\label{tab:summary}
\end{table}
The results in Table~\ref{tab:summary} provide useful insights into the relative predictive performance of the different methods in different simulation scenarios. In particular, Random Forest with differentiated inputs proves to be the best performing method, achieving the lowest median value across different complexities, including scenarios with jumps, random walks or a combination of both. While XGBoost is competitive, it tends to have a slightly higher median value under these conditions. Traditional time series methods such as ARIMA, SARIMA and TBATS consistently show robust and similar performance.

%
 \section{Real-World Data Example}
 \label{sec:DataEx}
\begin{figure}[h]
 \centering
 \includegraphics[width=0.9\textwidth]{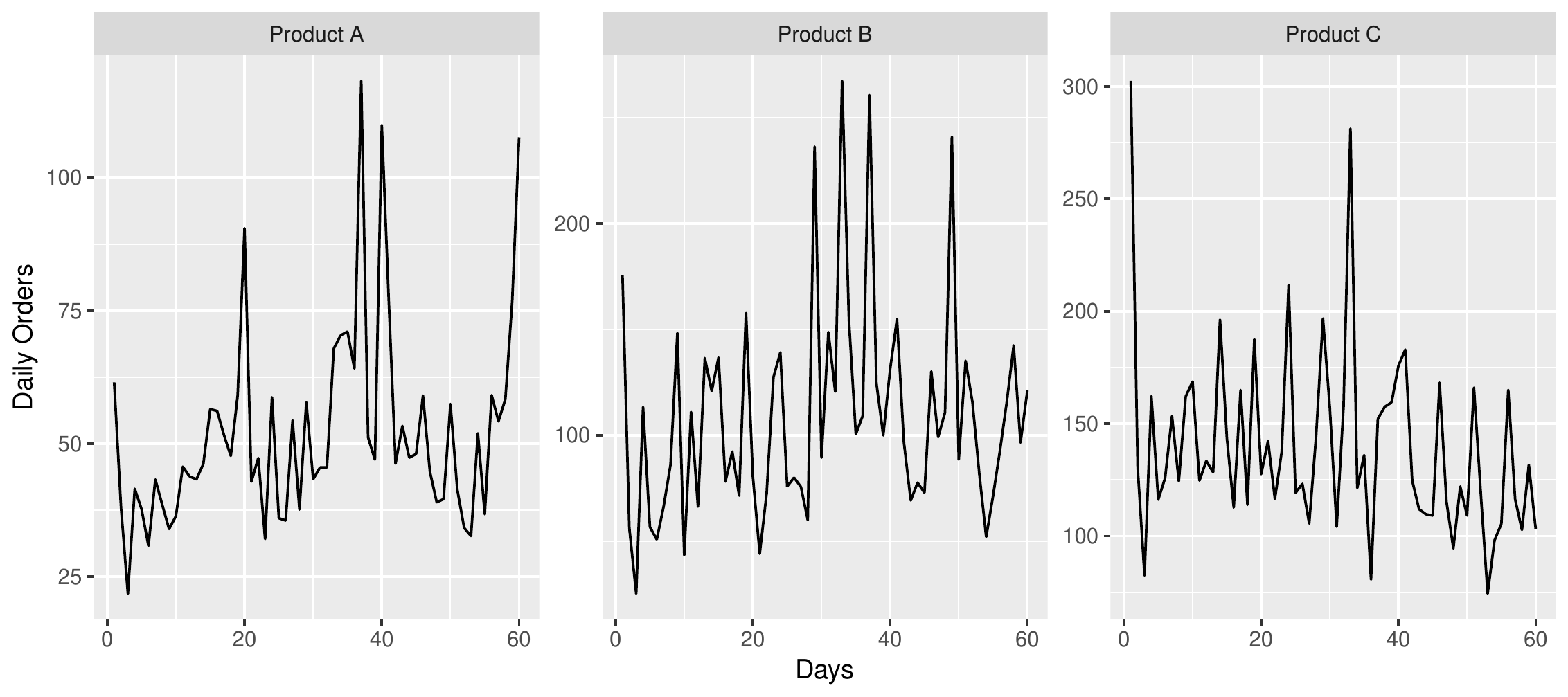}
 \caption{Daily orders of a Brazilian logistics company separated by the different products.}
 \label{fig:DataEx}
\end{figure}

%
As explained at the onset, there is a lack of freely available and good documented data sets in logistics research. We therefore use a rather simple real-world data example for illustration. The data set contains daily demand orders from a Brazilian logistics company \citep{miscDaily} and was sourced from the UCI Machine Learning Repository \citep{Dua2019}. Covering a span of $60$ consecutive days, the data set 
consists of three time series that capture orders for products A, B, and C. Figure~\ref{fig:DataEx} shows the corresponding time series in which specific shocks in the data can be identified. This observation puts us in a similar setting to the simulation study where the DGP was overlaid with a Poisson process. Given this context, it is of interest to evaluate whether the robust performance of (differentiated) machine learning algorithms observed in the simulation study is also apparent in for this dataset.

 The machine learning algorithms adhere to the hyperparameters outlined in Section~\ref{sec:Design}, with a sliding window size of eight, as informed by insights from our simulation study. We use the first 50 observations to train all methods and the last ten observations to test the performance via time series cross validation \cite[Chapter~5.10]{hyndman2018forecasting}. The MSE and MAPE are again used as evaluation measures. The summarized results are presented in Table~\ref{tab:DataEx}. Note that the results of SARIMA and ARIMA are identical due to the absence of seasonality and are therefore combined into one method. 
\begingroup
\centering
\fontsize{8}{12}\selectfont
\setlength{\tabcolsep}{7pt} 
\begin{table}[h!]
\caption{Mean MAPE and MSE of the methods considered in Section \ref{sec:Methods} using daily demand order data set. }
\resizebox{0.85\textwidth}{!}{%
 \hspace{3.5cm}\begin{tabular}{l@{\hspace{13pt}}ccc@{\hspace{13pt}}ccc}
\toprule
 & \multicolumn{3}{c}{MAPE\hbox{\hspace{9pt}}} & \multicolumn{3}{c}{MSE\hbox{\hspace{9pt}}} \\
 \cmidrule(l{1.9pt}r{11pt}){2-4} 
 \cmidrule(l{1.9pt}r{11pt}){5-7} 
Method&Prod. A& Prod. B&Prod. C&Prod. A&Prod. B&Prod. C\\
 \midrule
 Random Forest& 24.30& 35.05&30.79&22.39&262.41&695.70\\
 Random Forest Diff&6.67&21.80&15.84&4.91&197.23&1.97\\
 XGBoost&25.06&41.62&19.51&22.34&376.62&147.20\\
 XGBoost Diff&10.70&37.98&27.15&13.10&841.56&41.00\\
 (S)ARIMA&28.57&49.30&33.56&29.48&1,142.14&655.88\\
 TBATS&28.37&36.17&33.56&43.14&446.18&663.78\\
 Naive&33.18&30.71&30.59&25.10&194.21&82.03\\

\bottomrule
\end{tabular} 
}\label{tab:DataEx}
\end{table}
\endgroup

The results show that the performance of the forecasting methods is different in the various product categories. In general, the machine learning algorithms deliver consistently better results than the traditional time series methods. This is in line with our simulation study, where ML methods showed better performances when additional complexities were present. 
Random Forest with differentiation performed best for all three time series and evaluation measures, again confirming the results obtained in the simulation study for such settings. It should be noted that the introduction of differentiation is beneficial for Random Forest in all predictions. For XGBoost, however, performance on product A improves significantly when differenced data is used, but in the other two time series differentiation leads to worse forecasting performance.\\

%
 \section{Summary, Discussion and Outlook}
 \label{sec:Conclusion}
%

\paragraph{Summary with Higlights} 
The main objective of this simulation study was to perform a one-step comparative analysis of prediction accuracy and evaluate the performance of tree-based machine learning and time series approaches that are typically used in data-driven logistics. Through a comprehensive investigation of different data generating processes, queueing models, and additional complexities, we aimed to determine each method's inherent strengths and limitations. Our analysis included conventional time series methods, including (seasonal) ARIMA models and TBATS, as well as machine learning methods such as Random Forest and XGBoost. In addition, we investigated the impact of data differencing on the performance of the two latter algorithms. 
The key findings from our study are as follows:
\begin{itemize}
 \item The out-of-the-box Random Forest emerged as the ML benchmark method.
 \item Training on differentiated time series can significantly improve the ML resilience.
 \item ML models are more robust with respect to additional (nonlinear) complexity, settings in which they outperformed statistical time series approaches.
 \item In all other settings, the time series approaches were at least competitive or even performed better. 
\end{itemize}

\paragraph{Detailed Discussion and Outlook} In our study, the Random Forest approach performed consistently better in all simulation settings than the XGBoost approaches. It is worth noting that no hyperparameter tuning was made in our study. Random Forests are known to be robust to hyperparameter settings and often perform well with default values \citep{probst2019, fernandez2014}. This robustness can be a crucial factor contributing to their superior performance compared to XGBoost. Applying techniques such as Bayesian Optimization or more simple grid or random search for hyperparameter tuning could change this observation and should be investigated in future studies. Regarding the effect of data differentiation on the performance of the two machine learning methods, we observed similar patterns. Differentiation improved performance, especially in queueing scenarios and situations where additional complexity was introduced into the data generation process. Without additional complexity, differentiation showed minimal impact, with the performance of both methods deteriorating slightly when the differentiated data was used, except for very linear data generation processes. Here, only a slight improvement was observed. This suggests that differentiation plays a crucial role in improving the resilience of machine learning methods, especially Random Forests when the data is overlaid with additional noise like a random walk.
When comparing the performance of the different time series approaches, we found subtle differences between them. ARIMA and SARIMA showed relatively similar performance in all simulation settings under consideration. Their prediction accuracy was quite consistent without big differences in most situations. Comparing their performance with that of TBATS, the differences are also small and not substantial, suggesting that ARIMA, SARIMA and TBATS had comparable predictive power in our simulation settings.
The additional complexity induced, such as a jump process or random noise, significantly impacts the predictive power. Introducing a jump process leads to increased MSE values for all methods and settings, indicating a significant impact on prediction accuracy. In this scenario, all methods show consistent behavior with strong increasing MSE values for increasing time series lengths. When a noise process is introduced, a more nuanced pattern emerges. For the machine learning approaches, differentiating the data proves beneficial and improves the overall performance. The Random Forest approach with differenced data as input outperforms the other approaches in most scenarios, closely followed by all three time series approaches. A comparison between Random forests and the time series approaches shows different performance patterns in the different simulation environments. In queueing situations, where the underlying processes are often characterized by complicated dynamics, the Random Forest approach shows superior performance. Furthermore, a notable trend emerges in simulation settings where a Poisson process complements the data generating processes. In these cases, ML methods show improved performance, indicating robustness to the inherent complexity introduced by the Poisson process. The adaptability of ML models to capture and learn from nonlinear patterns may contribute to their effectiveness in scenarios with Poisson process or random walk overlays. However, it is essential to recognize that this beneficial performance of ML methods is not universal. In all other simulation settings, the Random Forest approaches perform comparable or slightly worse than all three time series approaches. In addition to the simulation study, our illustrative data analyses were conducted with a focus on one-step demand forecasting for different products of a logistics company. The results indicate that machine learning algorithms can improve the forecasting performance in this context. In particular, the machine learning methods perform better or equally well as the time series methods for most products.\\

In the context of data-driven logistics, our results underscore the importance of tailoring time series forecasting methods to the specific characteristics of data sets encountered in different logistics areas. The Random Forest approach, especially when using differentiated data as input, is recommended as an initial benchmark prediction tool, particularly for data sets with a lot of noise or complex patterns. The robustness of Random Forests, combined with their ability to achieve good results without extensive tuning of hyperparameters, makes them a pragmatic choice for various prediction scenarios. Conversely, in situations where interpretability is paramount (e.g., to gain understanding or trust of users in warehouses or decision makers in SCM) and the data exhibit clear patterns, traditional time series approaches remain a valuable and interpretable option. These approaches often come with faster runtimes and greater resource efficiency, which is also essential in the development of data-driven logistics, e.g. in case of resource constraints \citep{venkatapathy2015phynode,gouda2023grid}. As only one-step forecasts were considered, future simulation studies should investigate whether the same observations can be found for more step forecasting. Also, additional or hybrid methods must be investigated \citep{aladag2009forecasting, zhang2003time, smyl2020hybrid}. Another line of future research needs to compare the methods with respect to uncertainty quantification, i.e., point-wise or simultaneous prediction intervals and regions.

\bibliography{paper}

 \newpage
 \appendix

\section{Additional Simulation Results}
\begin{figure}[h]
 \centering
 \includegraphics[scale=0.8]{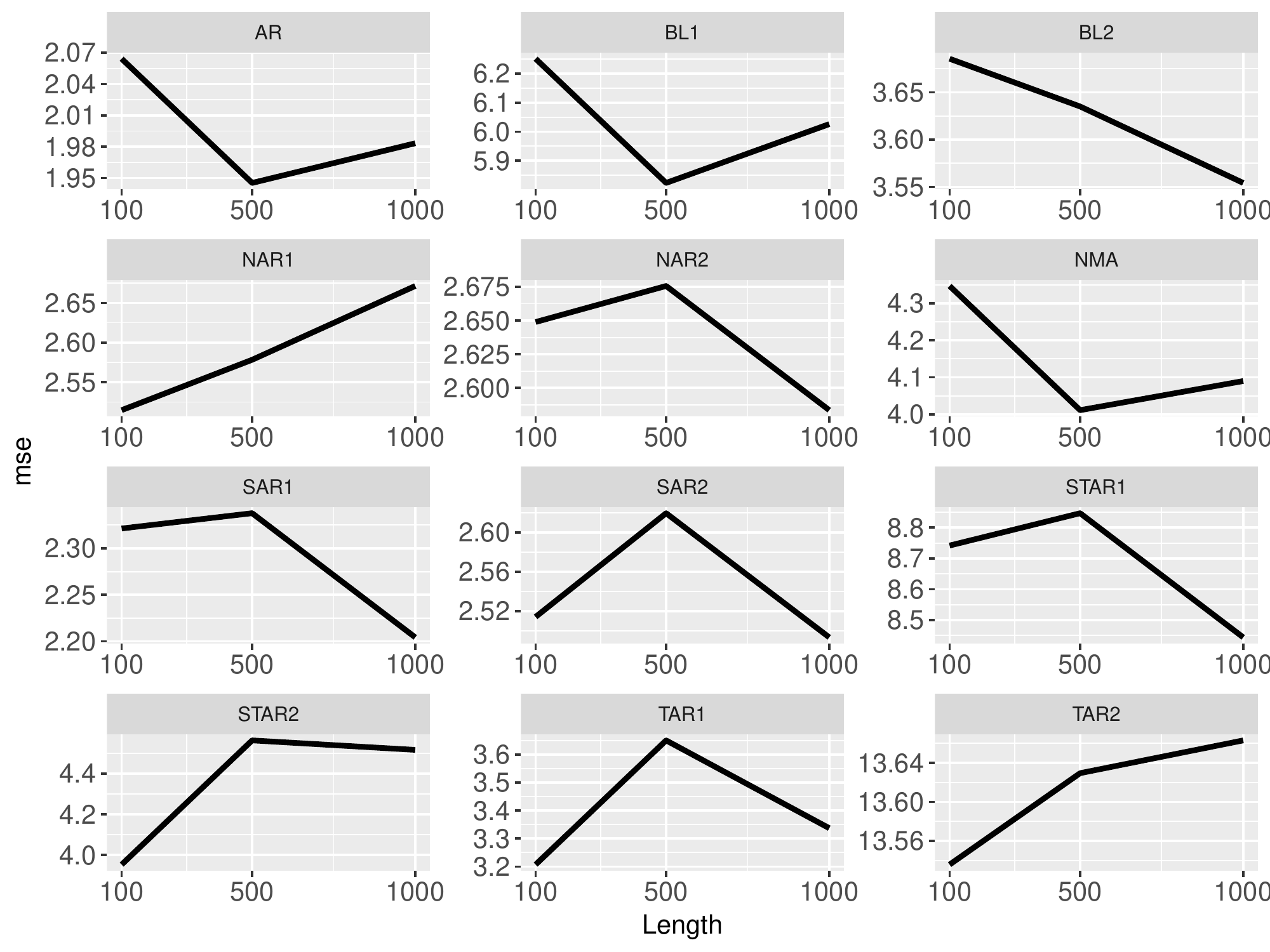}
 \caption{MSE of the naive approach for the different data generating processes.}
 \label{fig:naive}
\end{figure}
\subsection{Influence of Jump Process}
\begin{figure}
 \centering
 \includegraphics[width=0.9\textwidth]{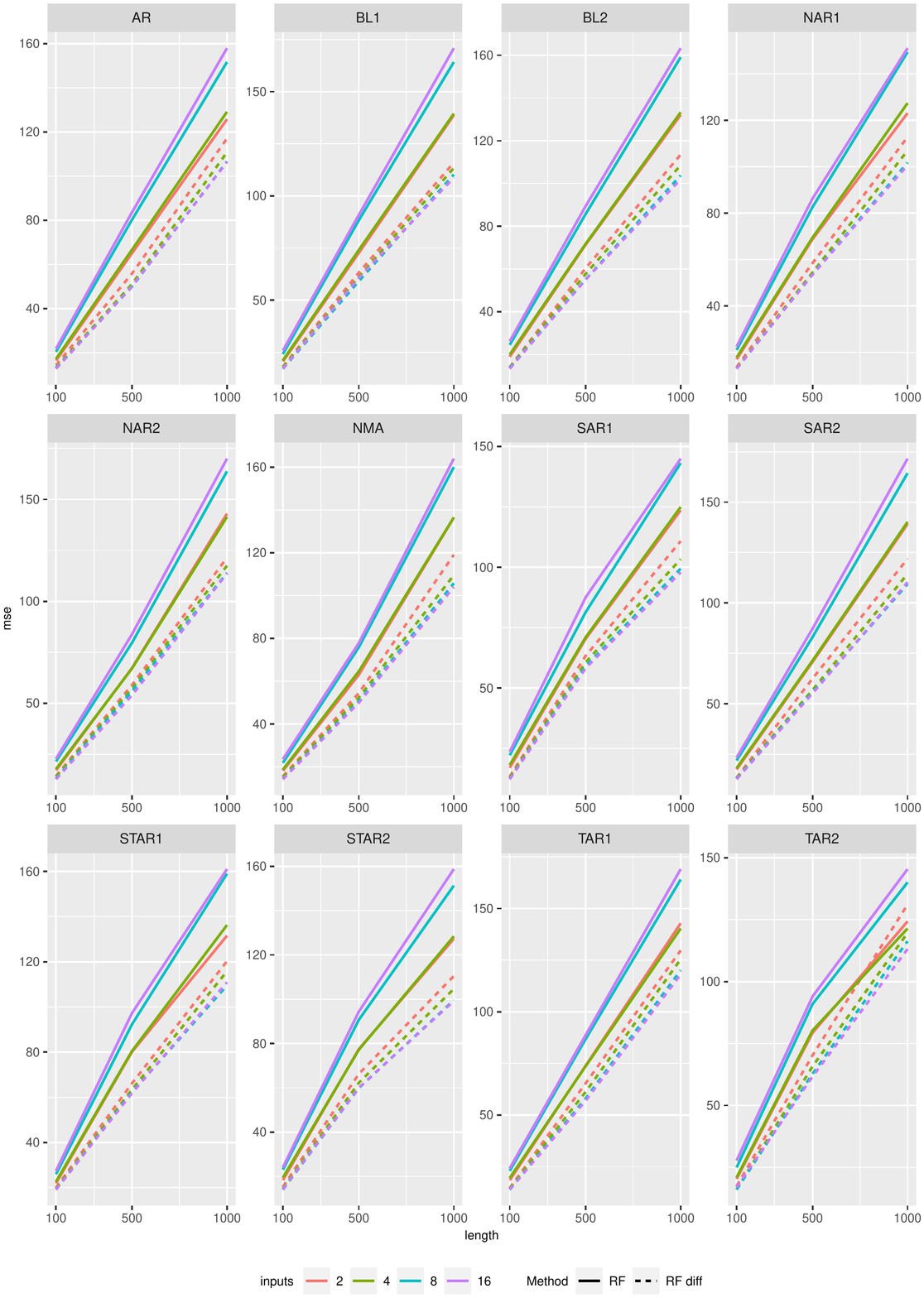}
 \caption{MSE values of all Random Forest approaches, sliding window sizes and data generating processes superposed by a compound Poisson process.}
 \label{fig:jump1}
\end{figure}

\begin{figure}
 \centering
 \includegraphics[width=0.9\textwidth]{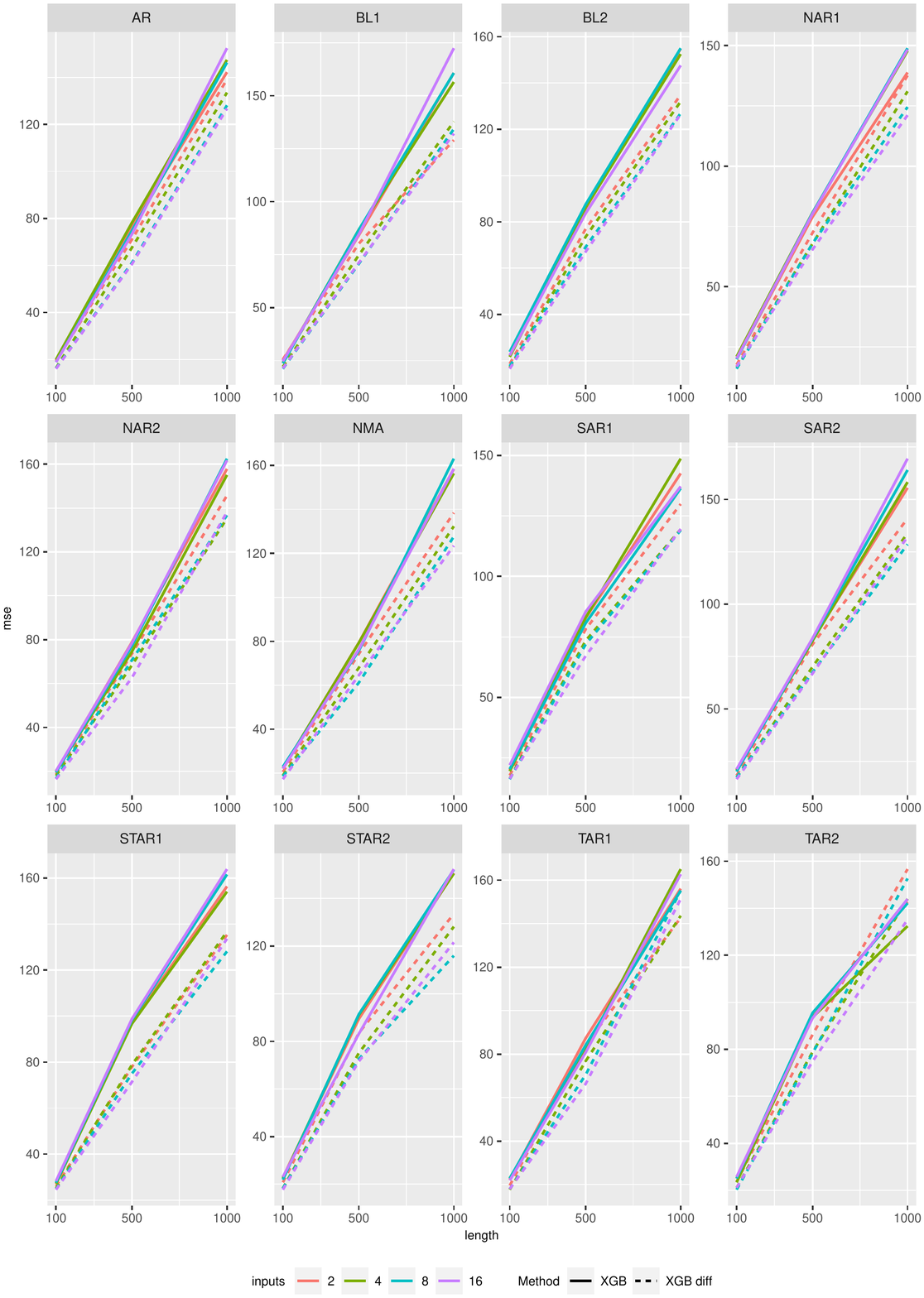}
 \caption{MSE values of all Random Forest approaches, sliding window sizes and data generating processes superposed by a compound Poisson process.}
 \label{fig:jump2}
\end{figure}

Figure \ref{fig:jump1} and \ref{fig:jump2} summarize the prediction results for all sliding window sizes and data generating processes using the ML methods. For both methods applied to differenced data, the performance is quite similar across the different windows sizes. However, a small difference in MSE values can be observed for the Random Forests, where a smaller window size slightly improves the prediction power.

\newpage

\subsection{Influence of Additional Noise}

\begin{figure}[h!]
 \centering
 \includegraphics[width=0.85\textwidth]{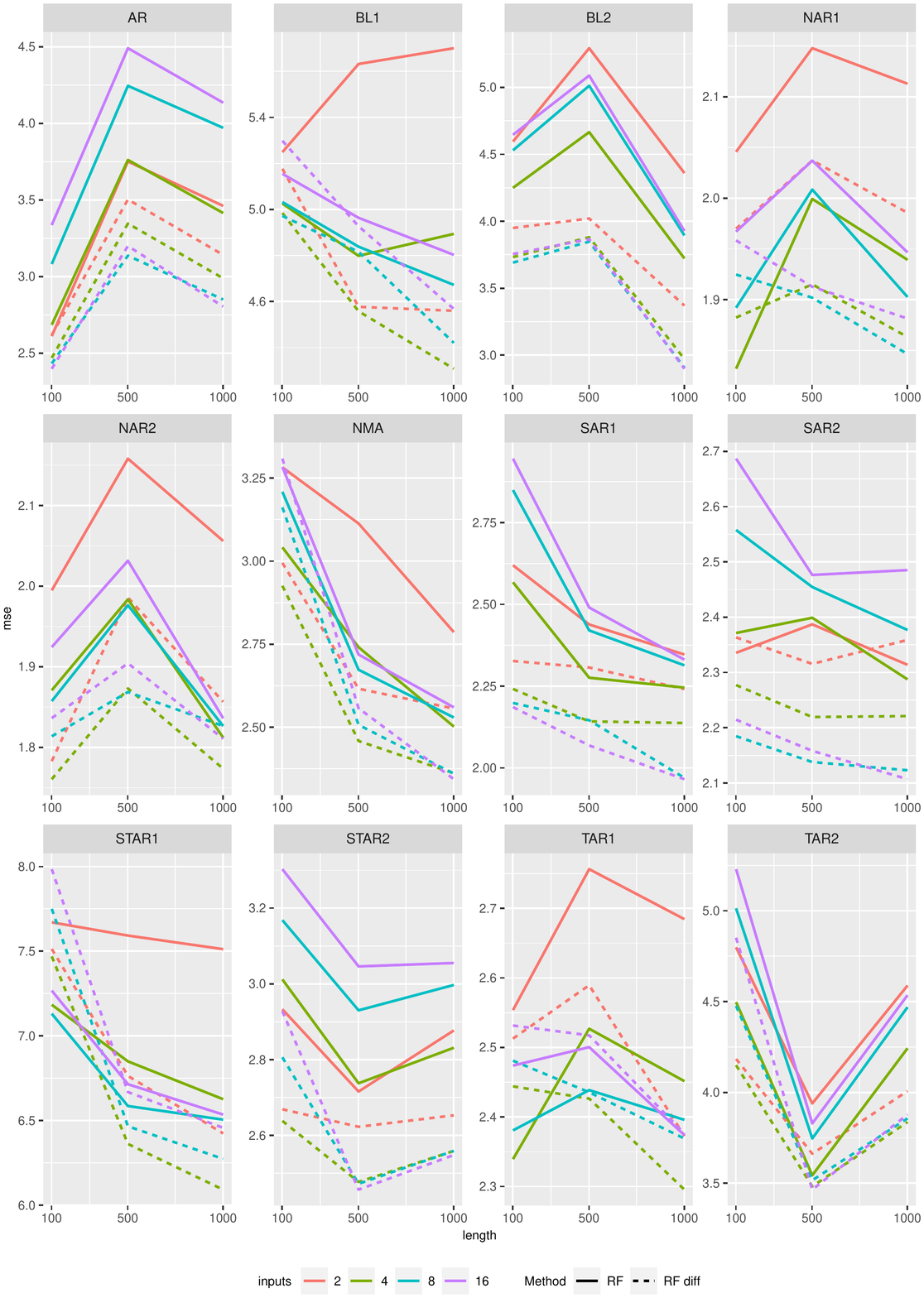}
 \caption{MSE values of all Random Forest approaches, sliding window sizes and data generating processes superposed by a random walk.}
 \label{fig:rw1}
\end{figure}

\begin{figure}[h!]
 \centering
 \includegraphics[width=0.87\textwidth]{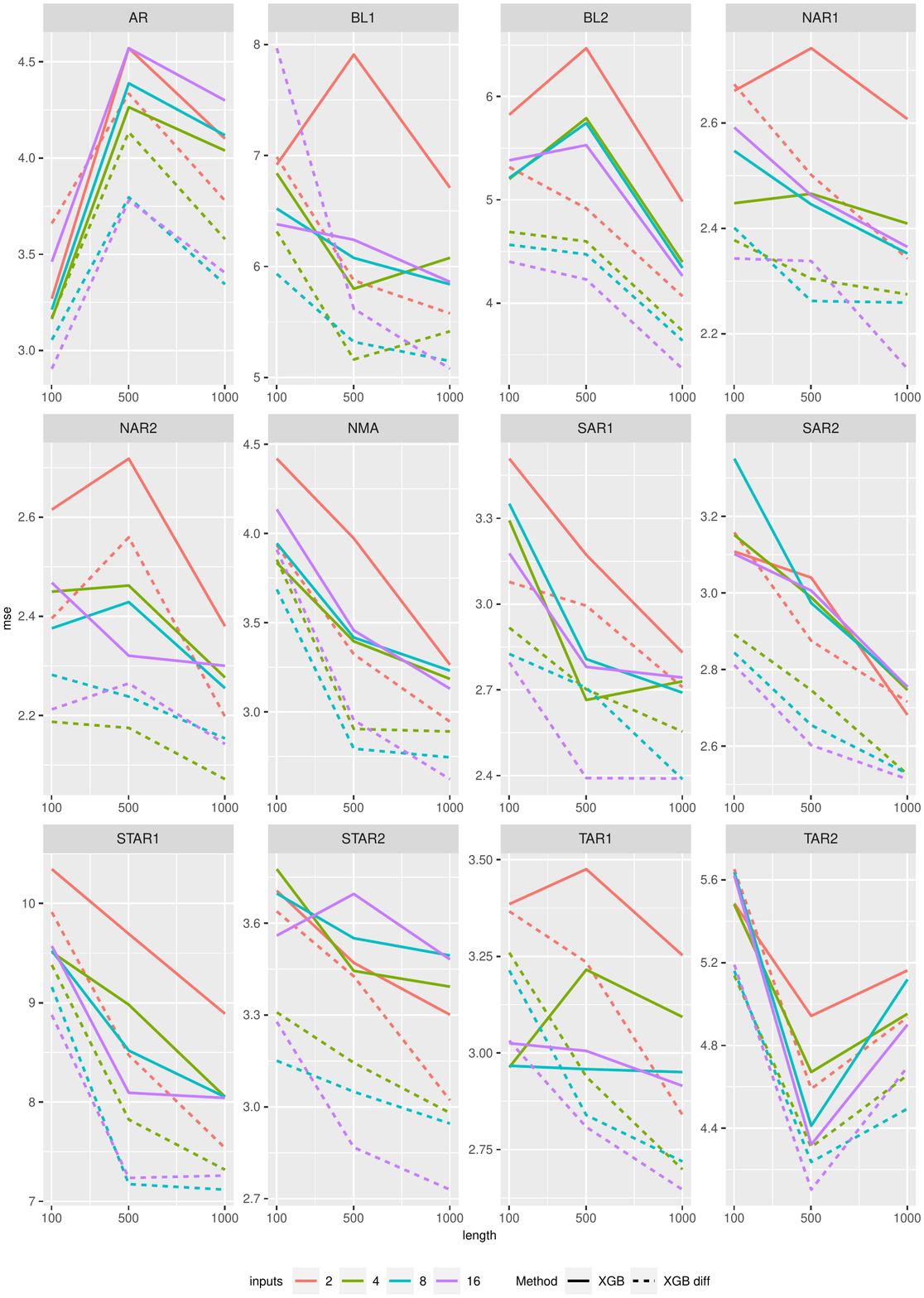}
 \caption{MSE values of all XGBoost approaches, sliding window sizes and data generating processes superposed by a random walk.}
 \label{fig:rw2}
\end{figure}

\newpage
\subsection{Influence of Additional Noise and Jump Process}

\begin{figure}[h!]
 \centering
 \includegraphics[width=0.85\textwidth]{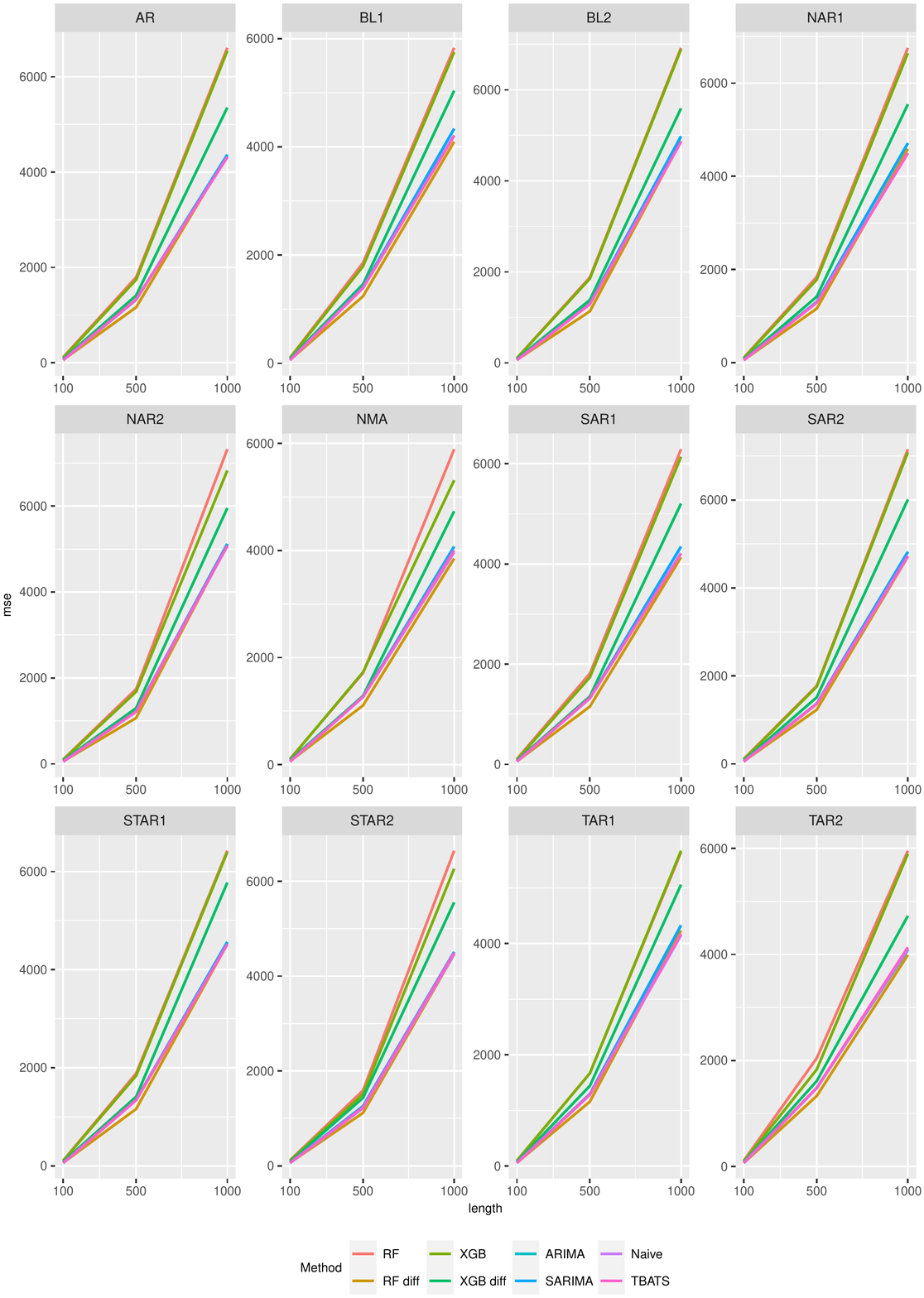}
 \caption{MSE of all methods and settings, where the data generating processes were superposed by a random walk and compound Poisson process.}
 \label{fig:both}
\end{figure}
\newpage
\begin{figure}[h!]
 \centering
 \includegraphics[width=0.85\textwidth]{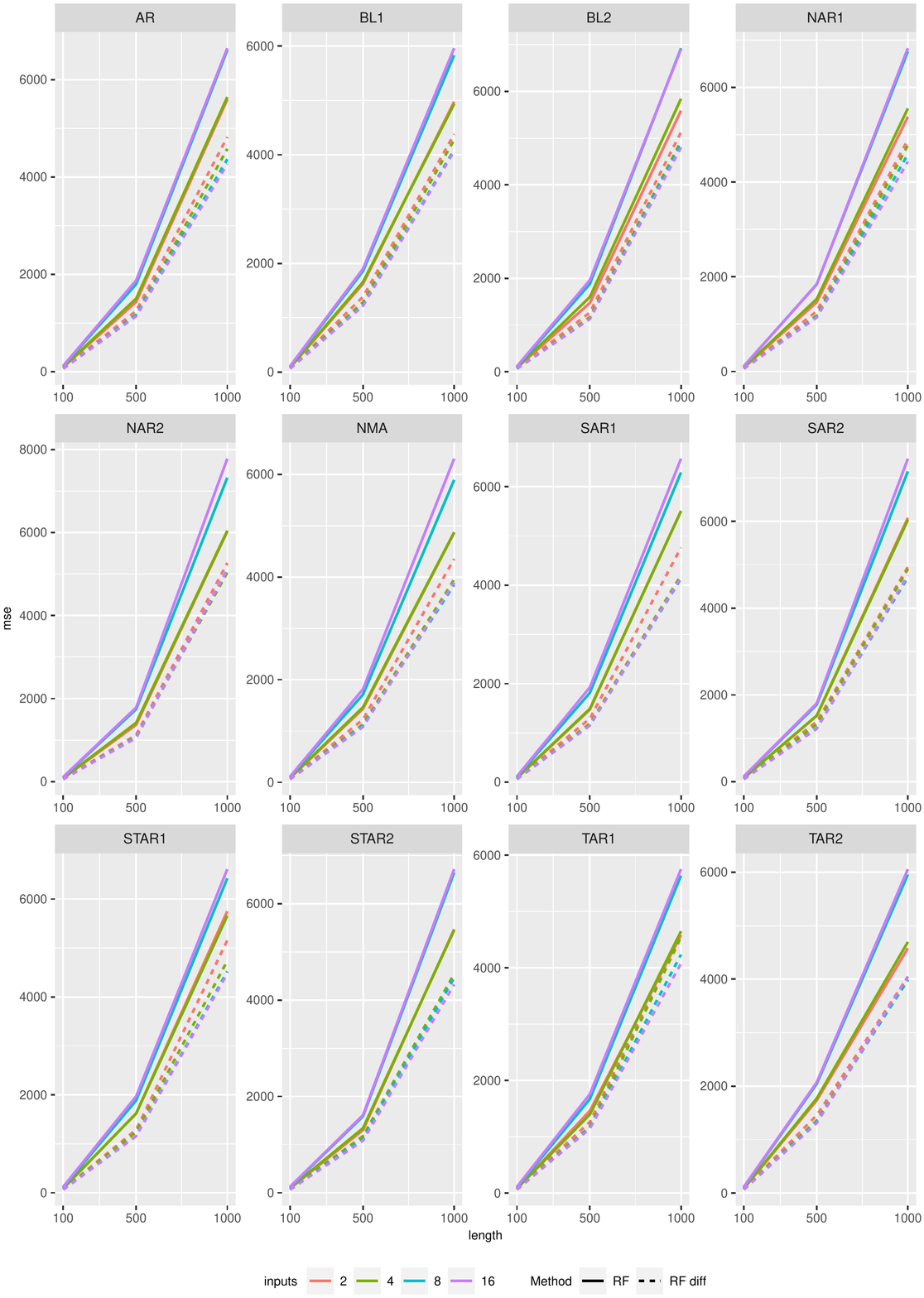}
 \caption{MSE of all Random Forest approaches, sliding window sizes and settings, where the data generating processes were superposed by a random walk and compound Poisson process.}
 \label{fig:both2}
\end{figure}
\newpage
\begin{figure}[h!]
 \centering
 \includegraphics[width=0.85\textwidth]{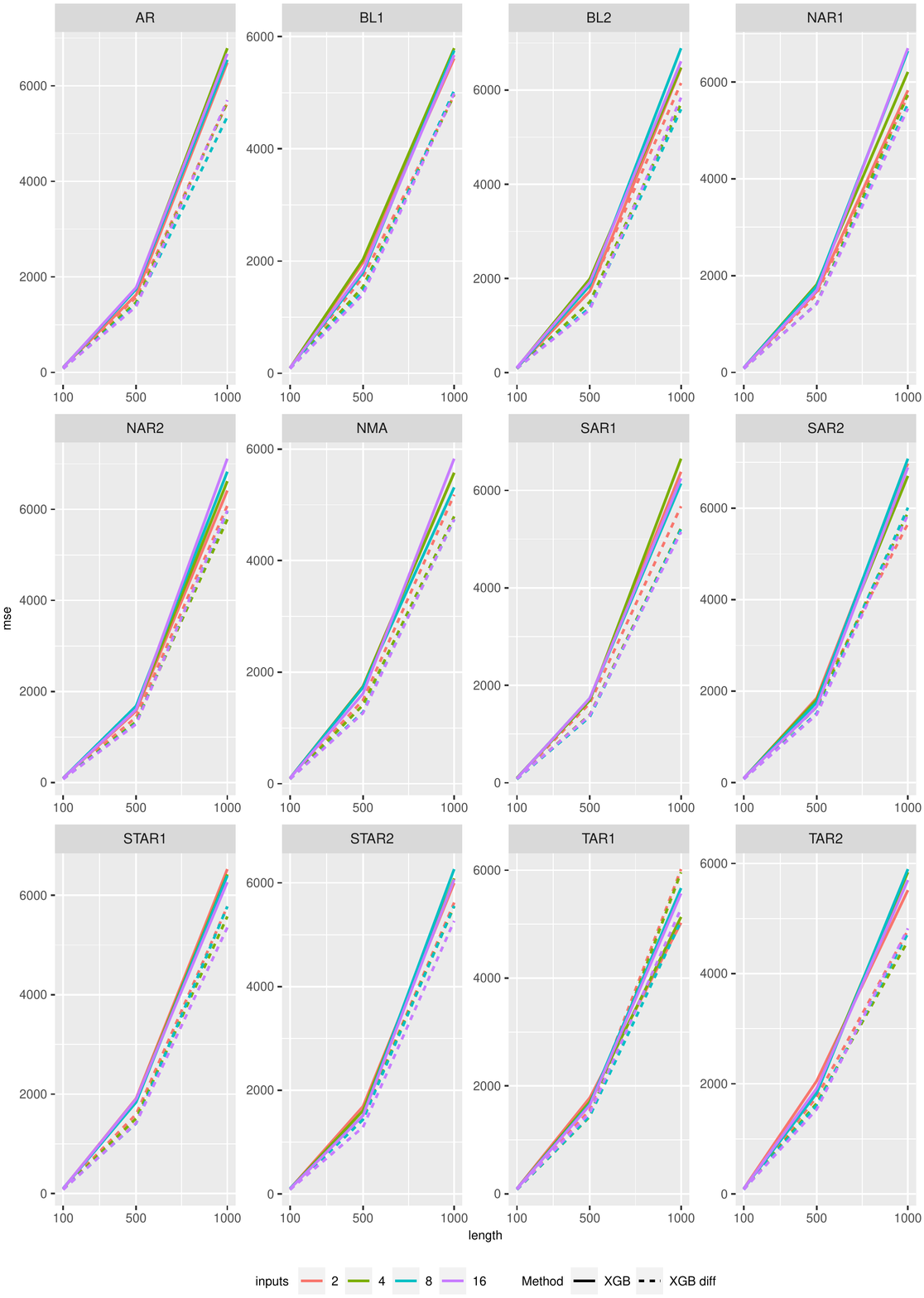}
 \caption{MSE of all XGBoost approaches, sliding window sizes and settings, where the data generating processes were superposed by a random walk and compound Poisson process.}
 \label{fig:both3}
\end{figure}

\newpage
\section{MAPE Results}
\begin{figure}[h!]
\centering
\includegraphics[width=0.85\linewidth]{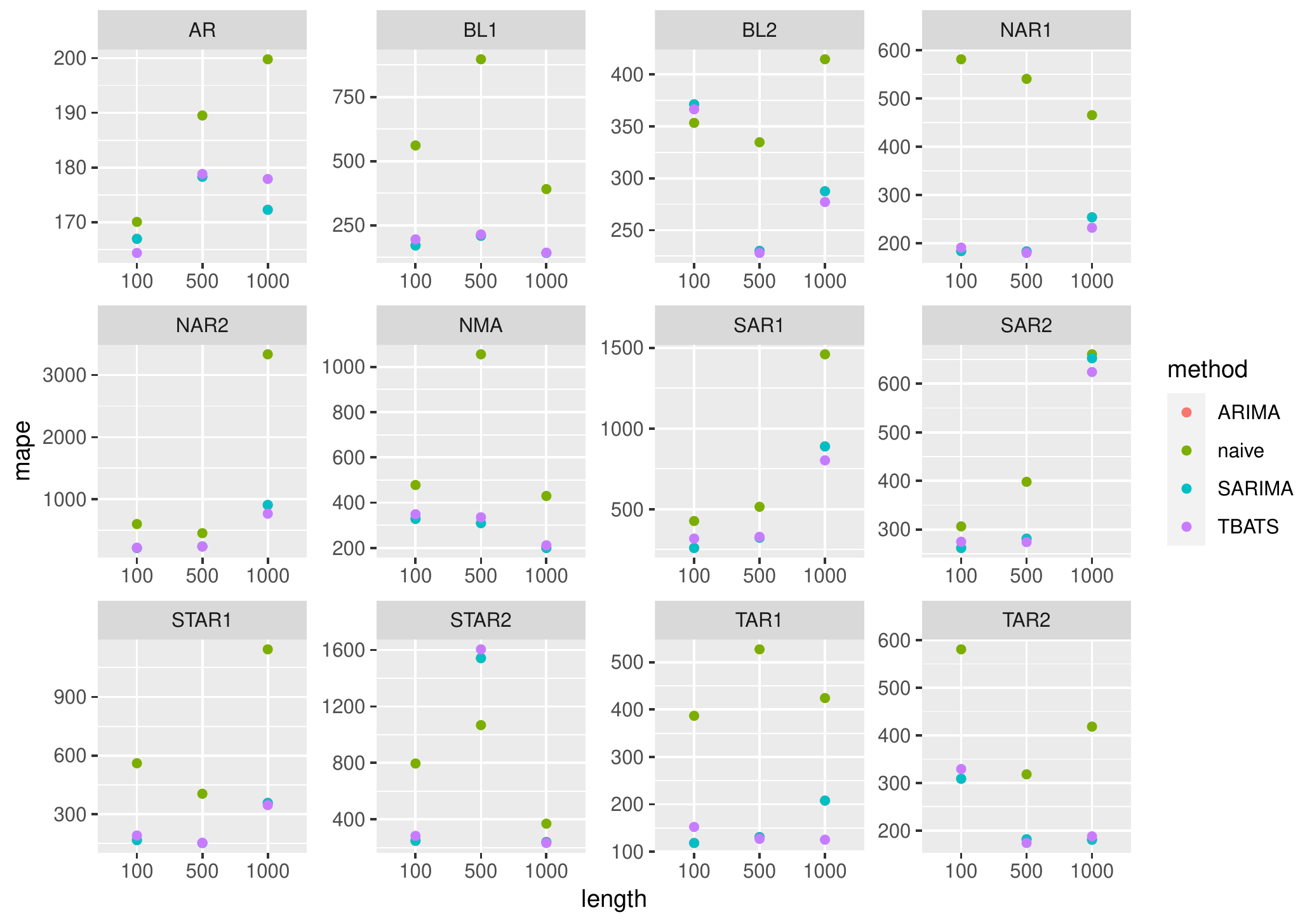}
\caption{MAPE of the time series approaches for the different data generating processes described in Table 1.}

\end{figure}

\begin{figure}[h!]
\centering
\includegraphics[width=0.85\linewidth]{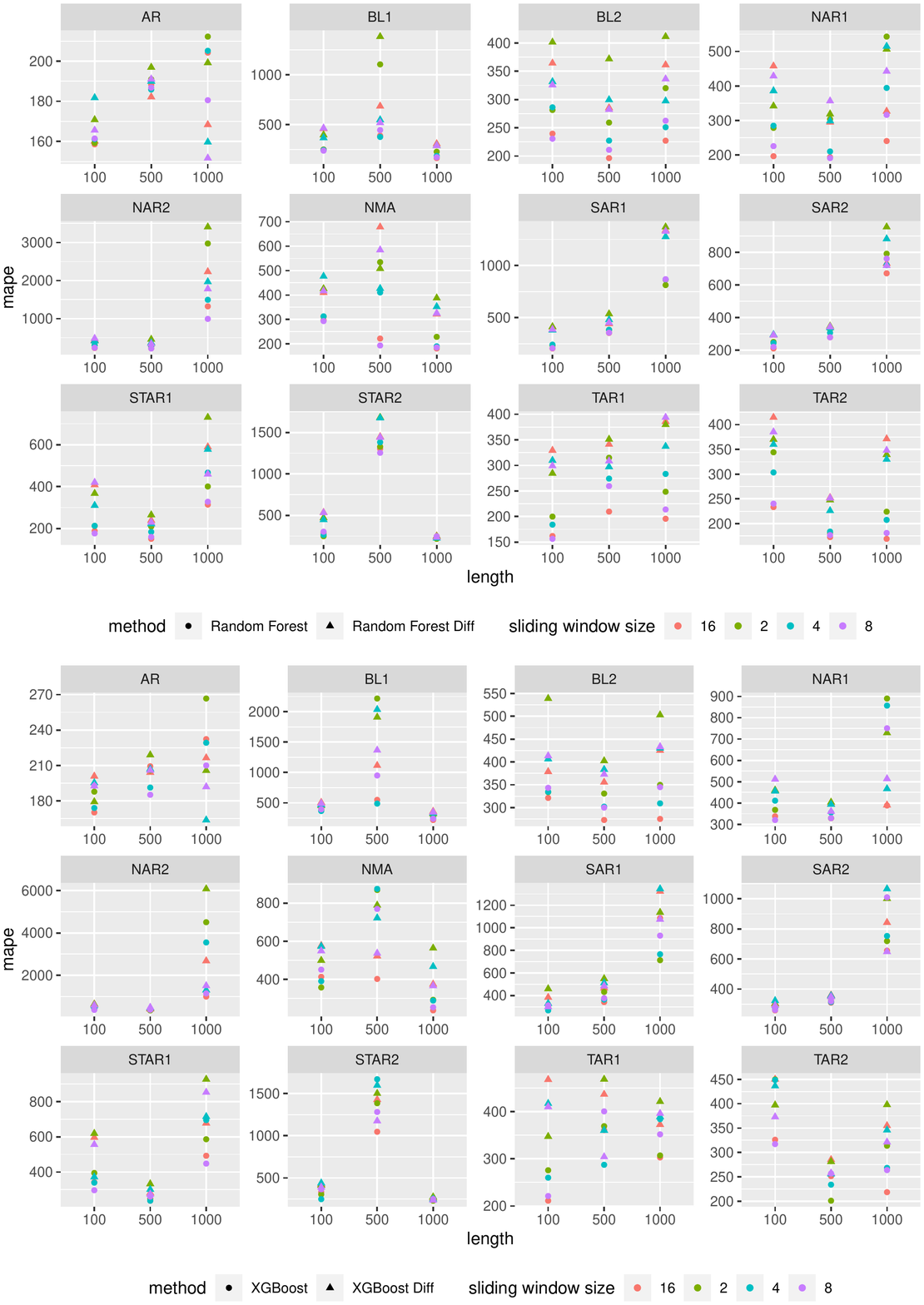}
\caption{MAPE of the Random Forest (above) and XGBoost (below) approaches for the different data generating processes described in Table 1 superposed by a compound Poisson process.}

\end{figure}

\begin{figure}[h!]
\centering
\includegraphics[width=0.8\linewidth]{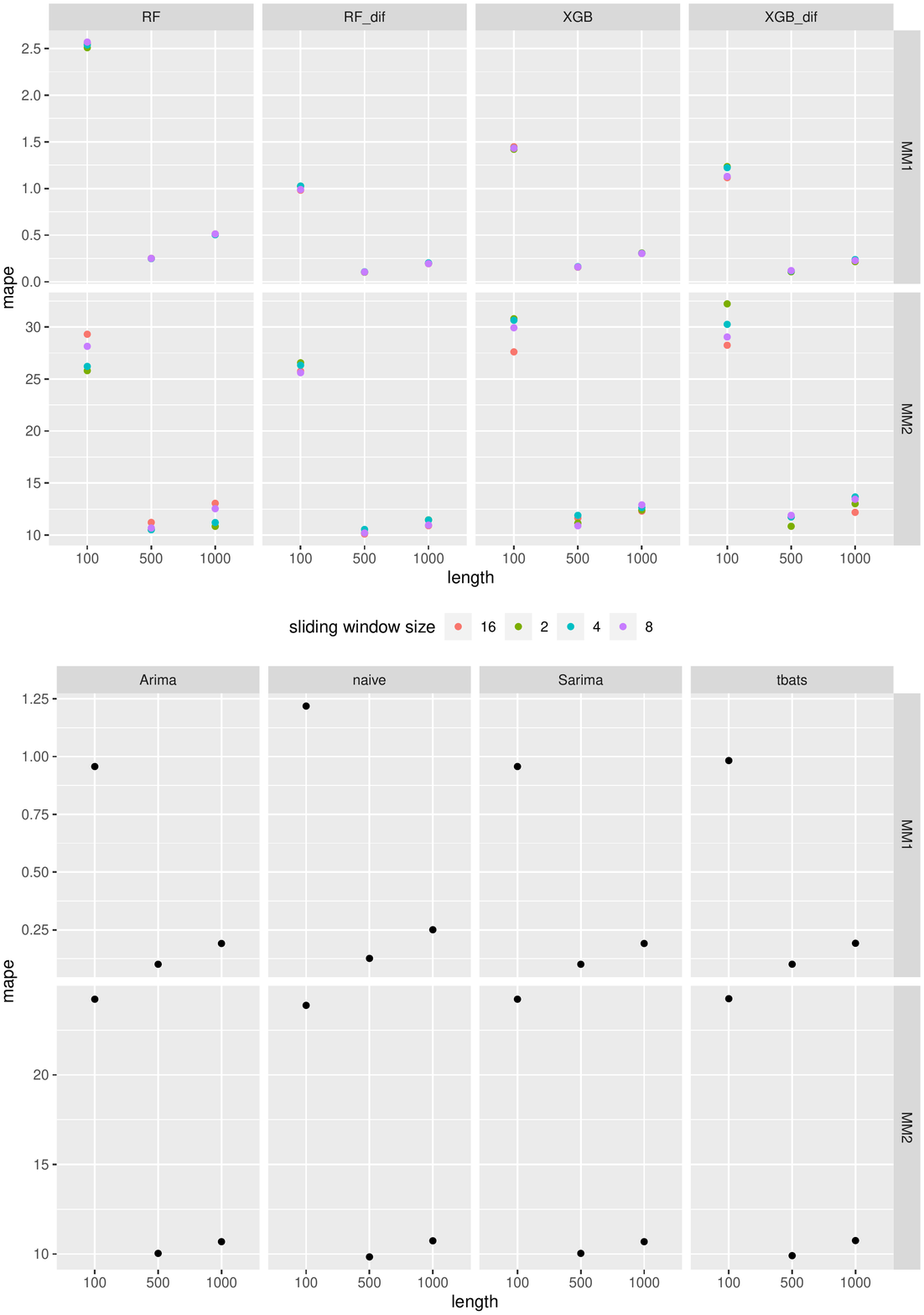}
\caption{MAPE of the machine learning algorithms (above) and time series approaches (below) for the M/M/1 and M/M/2 data generating process.}

\end{figure}

\begin{figure}[h!]
\centering
\includegraphics[width=0.85\linewidth]{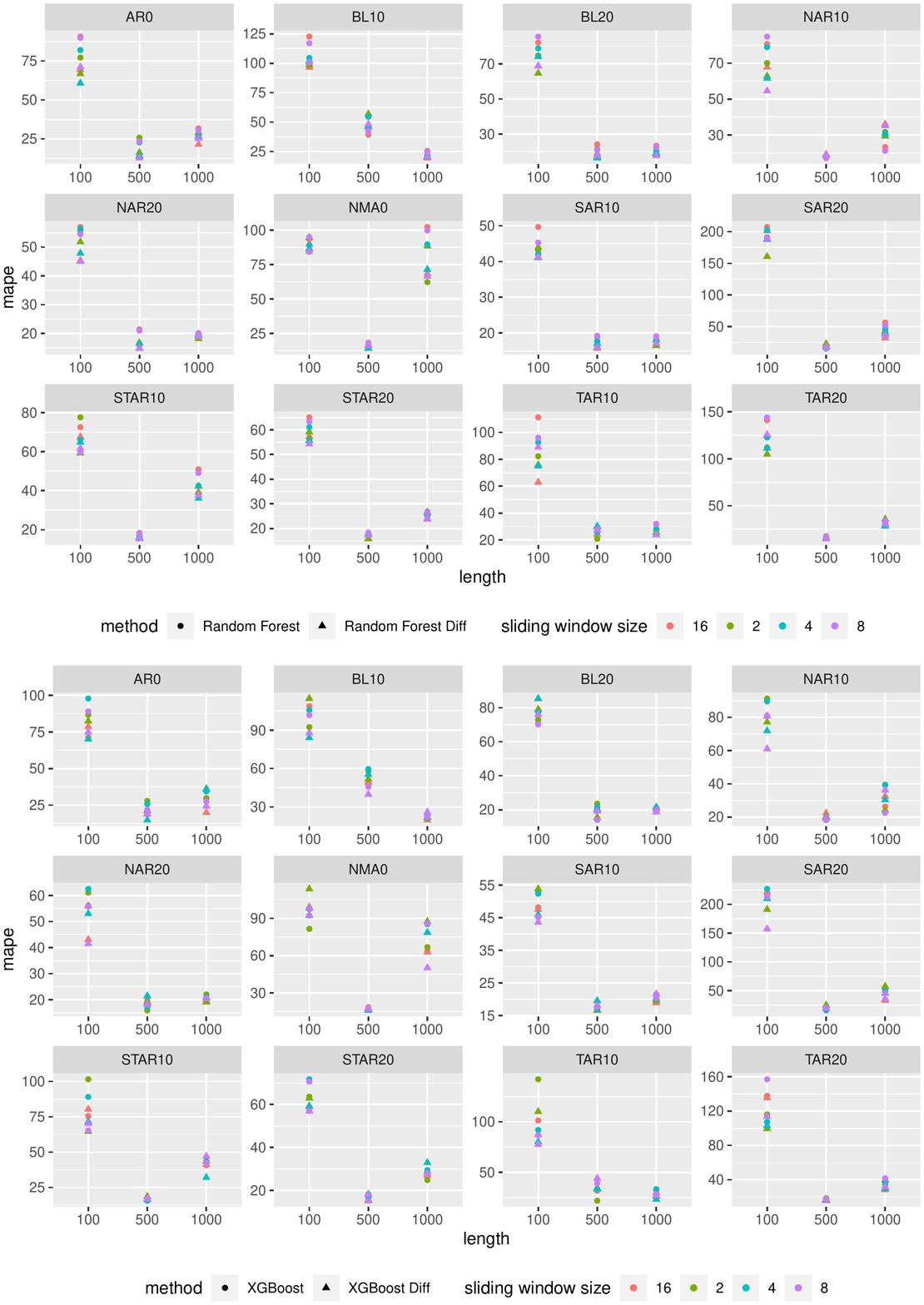}
\caption{MAPE of the Random Forest (above) and XGBoost (below) approaches for the different data generating processes described in Table 1 superposed by a compound Poisson process.}
\end{figure}

\begin{figure}[h!]
\centering
\includegraphics[width=0.85\linewidth]{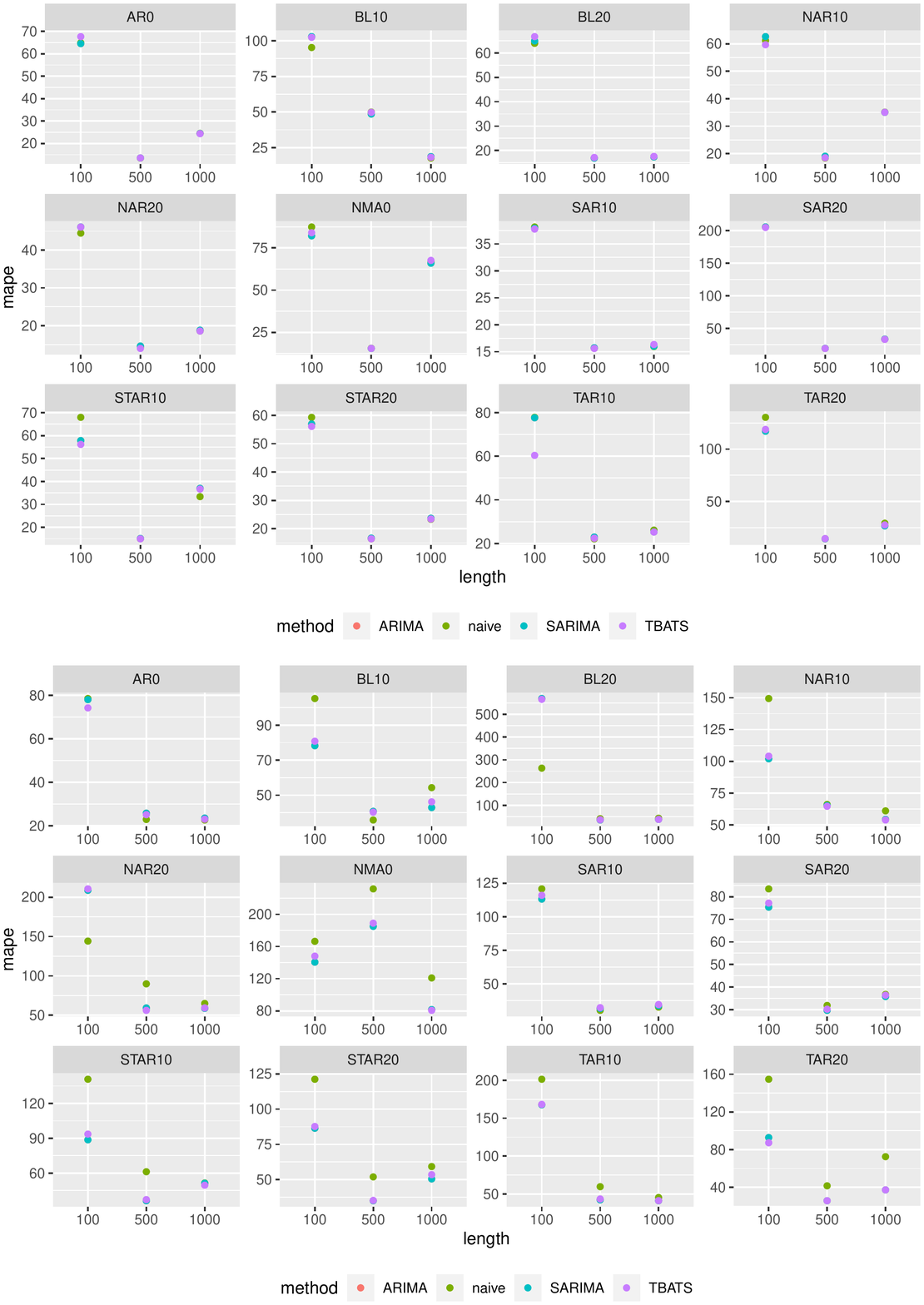}
\caption{MAPE of the time series approaches for the different data generating processes described in Table 1 superposed by a compound Poisson process (above) or a random walk (below).}

\end{figure}

\begin{figure}[h!]
\centering
\includegraphics[width=0.85\linewidth]{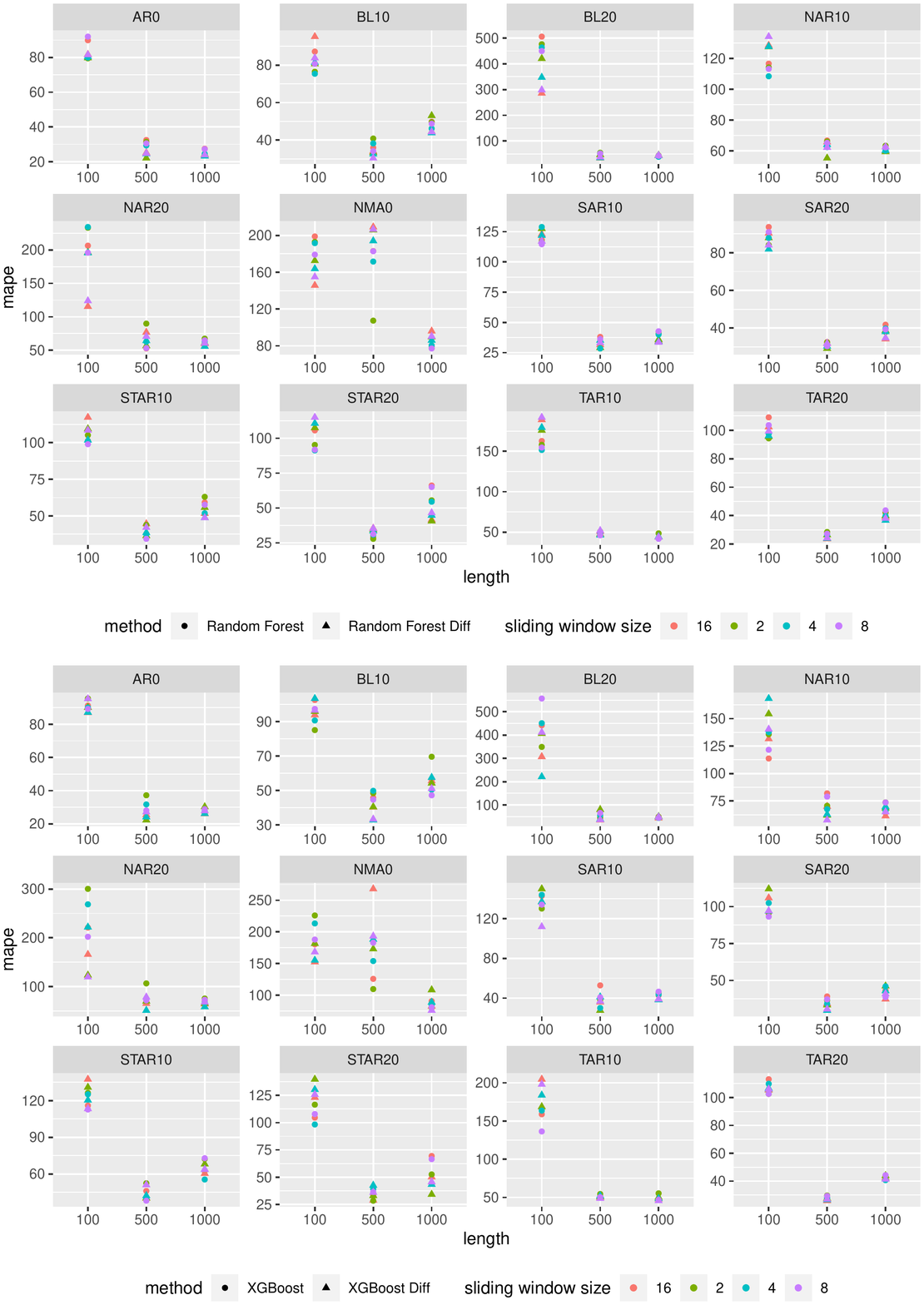}
\caption{MAPE of the Random Forest (above) and XGBoost (below) approaches for the different data generating processes described in Table 1 superposed by a random walk.}
\end{figure}

\begin{figure}[h!]
\centering
\includegraphics[width=0.85\linewidth]{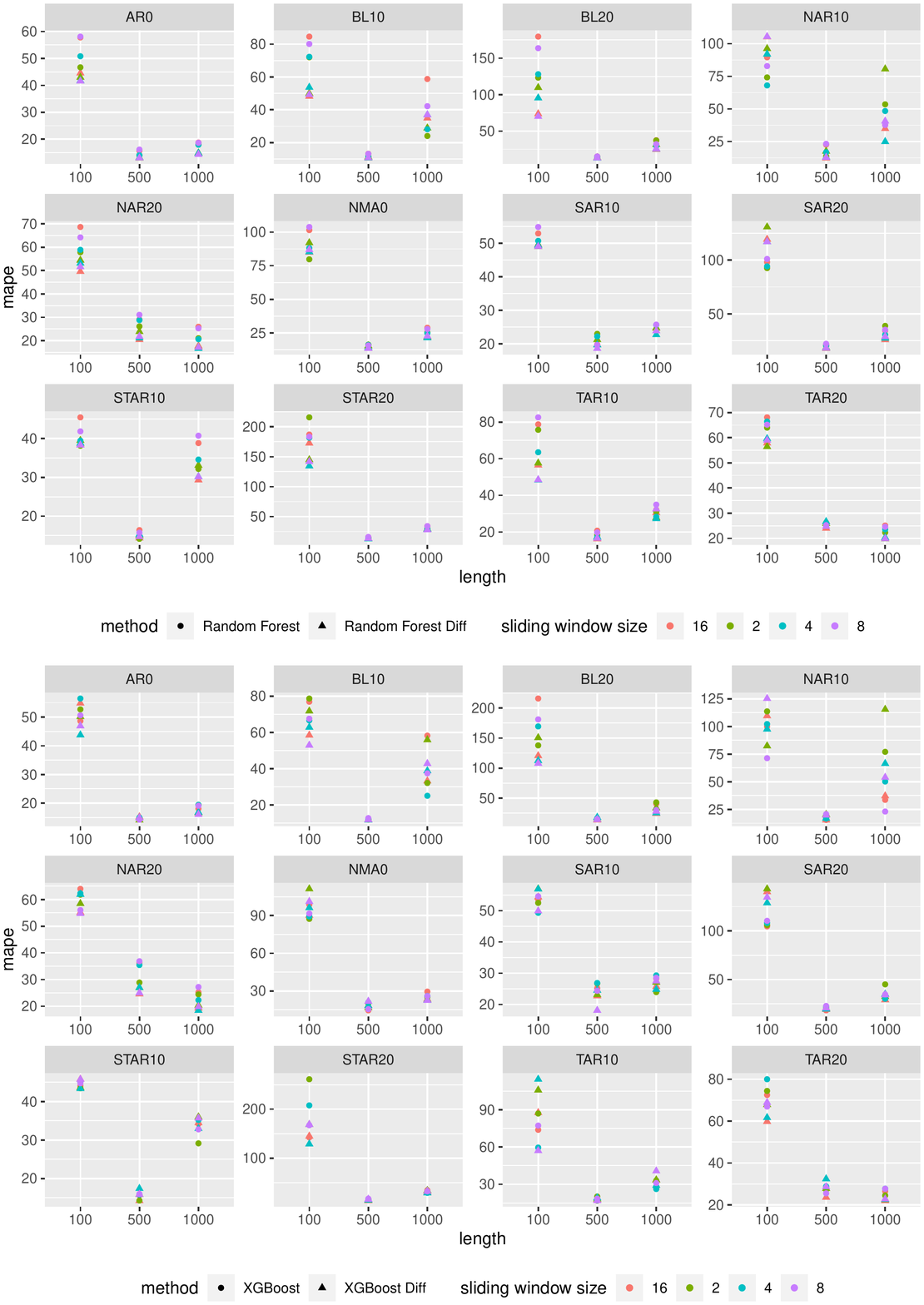}
\caption{MAPE of the Random Forest (above) and XGBoost (below) approaches for the different data generating processes described in Table 1 superposed by a compound Poisson process and a random walk.}
\end{figure}

\begin{figure}[h!]
\centering
\includegraphics[width=0.7\linewidth]{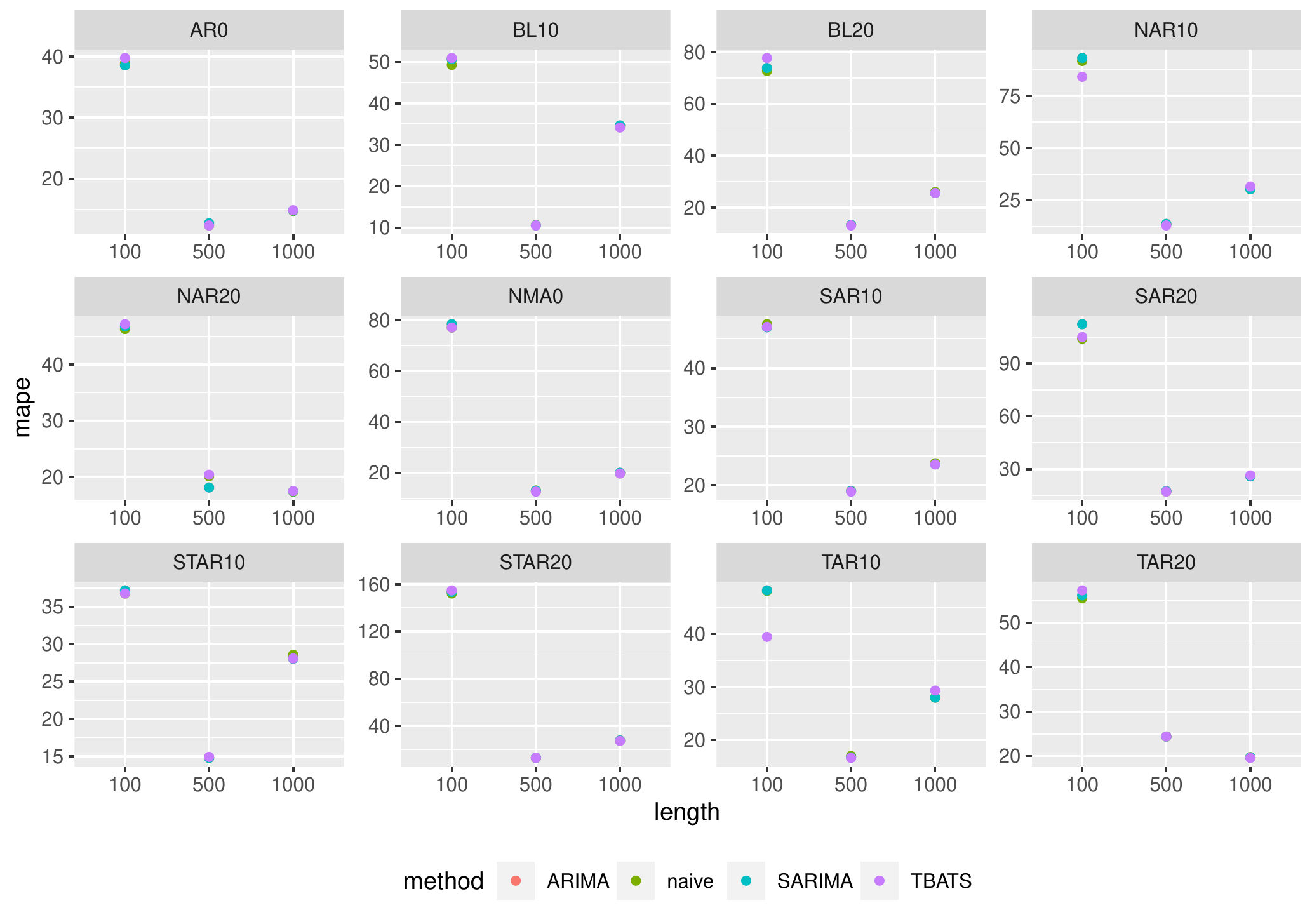}
\caption{MAPE of the time series approaches for the different data generating processes described in Table 1 superposed by a compound Poisson process and a random walk.}

\end{figure}

\end{document}